\documentclass[letterpaper]{article} 
\usepackage{aaai2027}
\usepackage[hyphens]{url}  
\usepackage{graphicx} 
\def\UrlFont{\rm}  
\usepackage{natbib}  
\usepackage{caption} 
\usepackage{algorithm}
\usepackage{algorithmic}
\usepackage{booktabs}
\usepackage{amsmath}
\usepackage{amssymb}
\nocopyright

\title{WebGrader: Training LLMs for Web Development with Self-Evolving Programmatic Grader}
\author{
    Boshui Chen\textsuperscript{\rm 1},
    Huiping Liu\textsuperscript{\rm 2},
    Shaolei Zhang\textsuperscript{\rm 3}\corresponding
}
\affiliations{
    \textsuperscript{\rm 1}Beijing Institute of Technology, China\\
    \textsuperscript{\rm 2}Beijing Jiaotong University, China\\
    \textsuperscript{\rm 3}Renmin University of China, China\\
    zhangshaolei98@ruc.edu.cn
}

\begin{document}
\maketitle

\begin{abstract}
Large language models increasingly generate complete websites from natural-language descriptions, and reinforcement learning has become a central approach to closing their remaining functional gap. This training regime is bottlenecked by reward design. Hand-authored browser scripts are executable yet costly to write for open-ended requirements, while VLM and GUI-agent graders scale but may issue verdicts before observing the decisive state. We propose \textbf{WebGrader}, a self-evolving programmatic grader that autonomously derives the required interaction flows from each website request, represents each flow as an executable Flow Contract, and uses its execution outcome as an RL reward. WebGrader materializes the generated project in a live browser, grounds target actions against the source code and live DOM, and collects visual, DOM, response, and persistent-state evidence along the same browser trajectory. A residual-driven offline loop then discovers reusable verifier skills, screens them on disjoint validation pages, and freezes the promoted skill graph before policy training. By separating test planning, action grounding, evidence collection, and semantic judgment, WebGrader issues a \textsc{Pass} verdict only after observing the requested transition. On WebGen-Bench, WebGrader trains an 8B policy to a \textbf{52.01\%} functional success rate, outperforming a matched appearance-plus-script reward by 7.88 points and surpassing o4-mini and DeepSeek-v4-flash. On WG-core-250, the policy reaches a Full Score of \textbf{44.953} and surpasses Qwen3-Coder-480B\footnote{Code and data: \url{https://github.com/boneykingofnone/WebGrader}.}.
\end{abstract}

\section{Introduction}

Large language models (LLMs) have made remarkable progress in code generation, advancing from function-level completion \cite{chen2021humaneval,austin2021mbpp} and repository-level program repair \cite{jimenez2023swebench} toward complete application construction.
Web development \cite{lu2025webgenbench,si2024design2code} has accordingly emerged as a demanding testbed because a generated website is at once an executable program, a rendered interface, and an interactive product whose correctness is revealed only after use.
Reinforcement learning with verifiable rewards offers a natural route to closing the remaining functional gap.
Still, its success in mathematics and algorithmic code \cite{shao2024deepseekmath} depends on an oracle that open-ended website requests rarely provide.
A prompt may describe buttons, forms, navigation, or persistent state, yet it does not come with the browser actions and state assertions needed to determine whether those behaviors actually work.

\begin{figure}[t]
\centering
\includegraphics[width=\columnwidth]{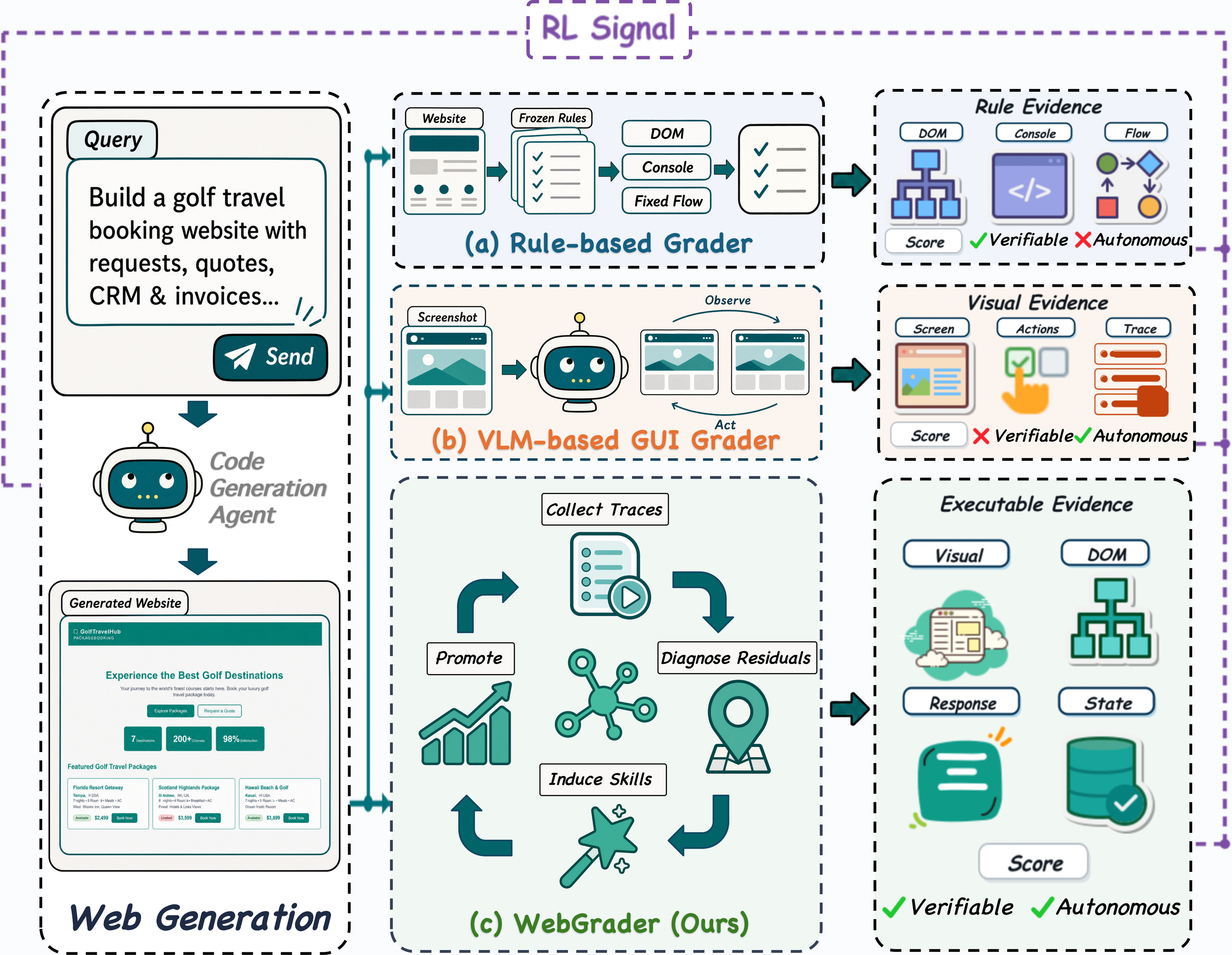}
\caption{WebGrader constructs and self-evolves an executable reward for web-development RL. It derives required interaction flows, grounds them as browser tests for each generated website, bases verdicts on execution evidence, and then freezes the grader.}
\label{fig:overview}
\end{figure}

Recent website-generation RL has begun to use screenshots and GUI-agent feedback as training signals, demonstrating the value of functional supervision \cite{lu2026webgenagent,jiang2026webgenr1}.
Existing web-development rewards provide such supervision in two main ways.
Hand-authored browser scripts, including task-specific web-agent evaluators \cite{zhou2023webarena,drouin2024workarena} and deterministic tests for interactive HTML \cite{wu2026htmlcure}, offer precise execution checks but require substantial per-task engineering and must be rewritten or adapted for new requirements and website structures.
Model-based GUI graders instead use a vision-language model to interact with the rendered website and readily scale to open-ended requests \cite{koh2024visualwebarena,he2024webvoyager,zheng2024seeact}.
Thus, script-based rewards are verifiable but not autonomous, whereas GUI graders are autonomous but not independently verifiable.
WebGrader combines both properties through autonomous, replayable browser tests grounded in execution evidence.

We therefore formulate open-ended web-development RL as a problem of \emph{executable reward construction}.
We first build a requirement-conditioned verifier that turns browser interactions into inspectable rewards, evolve it beyond its initial blind spots, and freeze it for policy training.
This requires grounding open-ended requirements in browser tests without accumulating a monolithic rule prompt.

To this end, we introduce \textbf{WebGrader}, a self-evolving programmatic grader for LLM-based web development.
WebGrader first derives the required interaction flows from the requirement alone, then grounds their Flow Contracts against the generated project and live DOM.
The resulting Playwright programs execute those flows and collect visual, DOM, response, and persistent-state evidence for per-flow judgments.
To improve the verifier itself, a neural-architecture-search-inspired offline loop~\cite{zoph2017nas} attributes residuals to planning, grounding, evidence, or judgment and mutates only the corresponding skills.
Promoted skills are organized into a routed SkillGraph with dependency, composition, and conflict relations, then fixed before reinforcement learning.
As shown in Figure~\ref{fig:overview}, WebGrader turns open-ended requirements into inspectable rewards while repairing recurrent verifier blind spots.

In summary, our key contributions are three-fold.
\begin{itemize}
    \item \textbf{Executable Reward Construction.} To the best of our knowledge, WebGrader is the first framework to autonomously turn open-ended requirements and generated code into browser-executable, evidence-grounded rewards for web-development RL.
    \item \textbf{Self-Evolving SkillGraph.} We develop a NAS-inspired loop that converts attributed verifier residuals into skill mutations, promotes them on disjoint validation pages, and composes them through a routed SkillGraph.
    \item \textbf{RL Supervision.} WebGrader trains an 8B policy that improves over a matched appearance-plus-script reward by 7.88 FSR points and surpasses o4-mini and DeepSeek-v4-flash on both WebGen-Bench and WG-core-250.
\end{itemize}

\section{Related Work}

\textbf{Verifiable Reward Construction.}\quad
Reinforcement learning with verifiable rewards is most direct when a test oracle already exists.
Code-generation benchmarks provide input and output examples or unit tests \cite{chen2021humaneval,austin2021mbpp,hendrycks2021apps}, expand them for stronger coverage \cite{liu2023evalplus}, or reuse repository tests for issue resolution \cite{jimenez2023swebench}.
When exact tests are unavailable, learned judges and task-conditioned rubrics provide broader but less directly checkable supervision \cite{liu2023geval,zheng2023llmjudge,kim2023prometheus,gunjal2026rubrics}.
Recent work instead constructs verification as part of the learning pipeline.
VerIF combines code and model-based verification, ReSyn synthesizes reasoning environments with code verifiers, EigenData co-generates tool-use trajectories and executable instance checkers, and EvolveCoder adversarially refines tests against candidate programs \cite{peng2025verif,he2026resyn,gao2026eigendata,ruan2026evolvecoder}.
These systems motivate constructed verification, but focus on instruction following, reasoning, or code, not requirement-specific browser flows over generated websites.

\textbf{Functional Web Evaluation.}\quad
Evaluation of generated websites has progressed from rendered similarity to executed interaction.
Design2Code measures visual fidelity \cite{si2024design2code}.
WebGen-Bench pairs website requirements with curated operation and expectation cases executed by a navigation agent \cite{lu2025webgenbench}, providing the UI-agent evaluation protocol used in our main experiments.
HTMLBench, introduced with HTMLCure, evaluates interactive HTML with deterministic browser programs \cite{wu2026htmlcure}.
We use its complementary test-based protocol through WG-core-250.
Related GUI testing systems evaluate repository-level playability with PlayEval and PlayTester, separate checklist construction from defect detection in WebTestBench, and diagnose exact target-state interactions in FineState-Bench \cite{peng2026playcoder,kong2026webtestbench,ji2026finestate}.
These benchmarks measure whether generated interfaces work, but do not themselves provide a self-evolving reward for training the website-generation policy.

\textbf{Functional Rewards for Web Development.}\quad
WebRenderBench optimizes layout and style consistency \cite{lai2025webrenderbench}, while ReLook scores rendered screenshots with a multimodal critic and penalizes invalid renders \cite{li2025relook}.
Moving toward functionality, WebGen-Agent uses screenshot and GUI-agent scores as step-level rewards \cite{lu2026webgenagent}.
WebGen-R1 combines structural and runtime feedback with visual supervision for end-to-end website-generation RL \cite{jiang2026webgenr1}, but does not compile open-ended requirements into programmatic tests with explicit postconditions.
An adjacent system, InfiniteWeb, automatically constructs functional websites and task-specific evaluators, but uses their verifiable rewards to train GUI agents that operate the generated environments, not models that generate the websites \cite{zhang2026infiniteweb}.
WebGrader differs in three respects.
It autonomously constructs requirement-conditioned Flow Contracts for each generated website, grounds verdicts in browser-executed evidence, and freezes a residual-evolved verifier for policy training.

\begin{figure*}[t]
\centering
\includegraphics[width=\textwidth]{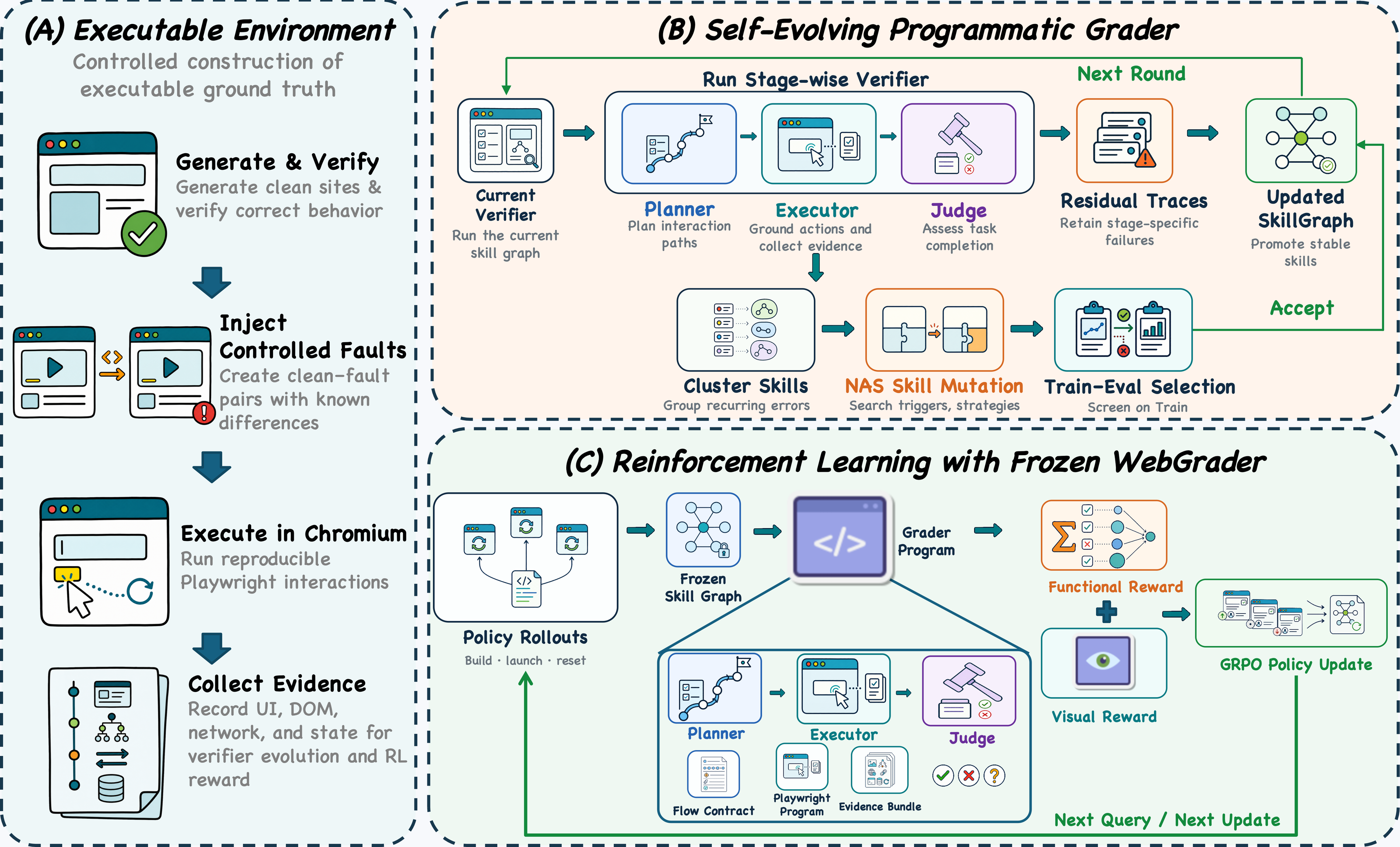}
\caption{End-to-end WebGrader pipeline. (a) Controlled clean/fault pairs provide executable supervision for offline verifier evolution. (b) Residuals are localized, mutated into verifier skills, and promoted through training/evaluation selection. (c) The promoted skill graph is frozen and routed over policy rollouts, producing evidence-grounded functional rewards for GRPO.}
\label{fig:full-pipeline}
\end{figure*}

\section{Method}

WebGrader is built on the principle that a grader should itself be evaluated and improved against quantified, verifiable outcomes before it is used as a reward.
We first construct an executable web environment that pairs verified clean applications with controlled faults and reference outcomes, making verifier variants measurable.
A base verifier is run in this environment, and its execution-grounded errors are abstracted into reusable skills and an initial SkillGraph.
A residual-driven, NAS-inspired process evolves the graph over multiple rounds by localizing new residuals, mutating stage-specific skills and relations, and promoting only candidates that improve validation performance.
The resulting graph and programmatic pipeline form WebGrader.
We then deploy WebGrader in reinforcement learning, where it constructs Flow Contracts, executes browser interactions, judges their evidence, and returns functional rewards for GRPO, completing the three-stage pipeline in Figure~\ref{fig:full-pipeline}.

\subsection{Executable Web Environment}

We instantiate this environment as \textbf{WebGen-Verifier-100}, using 100 public WebGen-Instruct training queries stratified across ten application categories.
Each query is paired with an accepted React/Vite application whose build, serve, render, and core requirement-level flows have been verified locally.
From every clean application, we materialize eight natural source-level single-fault variants by modifying localized business logic instead of injecting runtime markers.
A variant remains active only when the target flow is reachable, paired executions of clean and faulty sites expose the intended behavioral difference, the fault is isolated, and neither source contains explicit benchmark labels.
This process yields 100 clean applications and 778 active variants from 800 materialized faults.
We split at the page level into 60/20/20 Train/Eval/Test pages, containing 461/157/160 active faults and 282/95/93 active flows, so a clean application, all its variants, and their private references never cross splits.
Train is used for residual induction and mutation, Eval for candidate promotion, and Test only for reported verifier evaluation.
Test labels, fault specifications, expected transitions, and reference evidence remain hidden from the verifier and optimization, while Test outcomes never affect mutation, promotion, routing, early stopping, or hyperparameter selection.
The verifier, SFT, and RL query sets, containing 100, 600, and 600 requirements, are deliberately disjoint.

The environment provides five layers of ground truth comprising page and runtime metadata, Flow Contracts, isolated fault specifications, reference action-and-evidence traces, and scoring metadata.
A reference trace specifies an execution-semantic contract without prescribing a unique Playwright script, allowing different action sequences when they establish the same preconditions, execute the required target action, and collect equivalent before-and-after evidence.
The verifier observes only the requirement and normalized project.
Clean/fault labels, target-flow annotations, expected transitions, and reference evidence remain private to offline evaluation.
The environment rebuilds and launches every application in Chromium, resets its state, and executes interactions against the rendered interface.
It records screenshots, DOM states, responses, URL and network events, storage, refresh outcomes, and target-action reach.
These reference outcomes make clean acceptance, fault detection, precision, and evaluator-side uncertainty directly measurable.
Running a base verifier in this environment produces execution-grounded error traces that support SkillGraph induction and subsequent evolution.
The environment is used only for offline verifier construction and analysis.
RL rollouts are independently generated websites.

\subsection{Self-Evolving Programmatic Grader}

\textbf{Base Programmatic Verifier.}\quad
Given a requirement $q$, a verifier configuration $V$ first derives the required interaction flows $\mathcal{F}_{\mathrm{req}}(q)$ without consulting the generated website.
It then uses the generated project $x$, its source code, and the live DOM only to instantiate and ground every required flow as a Flow Contract
\begin{equation}
\begin{aligned}
\mathcal{C}(q,x)&=\{C_f(q,x)\}_{f\in\mathcal{F}_{\mathrm{req}}(q)},\\
C_f&=(P_f,A_f,a_f^\star,E_f,Q_f,M_f,w_f),
\end{aligned}
\label{eq:flow-contract}
\end{equation}
where $P_f$ contains preconditions and fixtures, $A_f$ is the interaction sequence, $a_f^\star$ is the target action, $E_f$ contains evidence checkpoints, $Q_f$ specifies postconditions, $M_f$ lists mandatory observations, and $w_f$ is the flow's criticality weight.
Grounding may add setup details, selectors, and alternative execution strategies, but it cannot remove a requirement-derived flow.
A required flow therefore remains in $\mathcal{C}(q,x)$ even when its control or behavior is absent from $x$.
It receives \textsc{Pass} only when the target action executes and the evidence supports $Q_f$.
Build or render failure, a missing control, an unexecutable target action, or a violated postcondition yields \textsc{Fail}.
\textsc{Inconclusive} is reserved for evaluator-side ambiguity, infrastructure exceptions, or malformed verifier output.

Verification exposes the intermediate artifacts
\begin{equation}
\begin{aligned}
\mathcal{F}_{\mathrm{req}}(q)&\xrightarrow{\mathrm{plan}(x)}\mathcal{C}
\xrightarrow{\mathrm{ground}(x)}\mathcal{S},\\
\mathcal{S}&\xrightarrow{\mathrm{execute}(x)}\mathcal{T}
\xrightarrow{\mathrm{judge}(q,\mathcal{C})}\mathbf{v},
\end{aligned}
\label{eq:verify}
\end{equation}
where $\mathcal{S}$ contains grounded Playwright scripts, $\mathcal{T}$ contains structured browser traces, and $\mathbf{v}$ contains per-flow verdicts.
Requirement analysis first derives the required flows from $q$ alone, the planner instantiates their Flow Contracts against $x$, the grounding stage resolves actions against the live DOM, and the evidence stage executes the scripts and records the visual, DOM, response, and persistent-state observations required by each contract.
A fixed evidence-conditioned LLM judge compares each trace with its postconditions and issues the semantic verdict.
Thus, \emph{programmatic} denotes executable tests and structured browser evidence with model-assisted planning, grounding, and judgment.

\textbf{Trace-Induced SkillGraph.}\quad
We run the base verifier $V_0$ on $\mathcal{D}_{\mathrm{train}}$ and retain its complete execution artifacts for every incorrect verdict or evaluator-caused inconclusive outcome.
An offline meta-evaluator clusters recurring error traces and abstracts them into reusable skills with activation triggers, localized stage transformations, evidence obligations, and dependency, composition, or conflict relations.
Together, they initialize a routed SkillGraph.

\textbf{Multi-Round Attribution and Mutation.}\quad
The initial SkillGraph is then self-evolved over multiple rounds.
At each round, the current verifier is replayed in the environment, and newly exposed errors become residuals.
Mirroring Eq.~(\ref{eq:verify}), an attribution function assigns each residual to a primary failure source as follows.
\begin{equation}
f_{\mathrm{attr}}(r) \in \mathcal{Z}
= \{z_{\mathrm{plan}},z_{\mathrm{ground}},
z_{\mathrm{evid}},z_{\mathrm{judge}}\}.
\label{eq:causes}
\end{equation}
Here the four sources denote an incomplete Flow Contract, failed browser-action grounding, missing evidence, and a verdict unsupported by the collected evidence, respectively.
For each attributed residual cluster, the offline meta-evaluator proposes a candidate mutation
$k=(\tau_k,\Delta_k,E_k,D_k)$, where $\tau_k$ specifies its trigger, $\Delta_k$ is a localized transformation of the affected stage, $E_k$ adds evidence obligations, and $D_k$ records dependencies, compositions, and conflicts.
Inspired by mutation and selection in neural architecture search~\cite{zoph2017nas}, each candidate targets one attributed source and applies a small, inspectable modification to the current verifier.
For example, a fixture skill creates missing prerequisites, an evidence skill adds before/after and refresh checkpoints, and a target-action skill prevents a success message from substituting for the requested interaction.
These skills modify verifier artifacts while leaving model weights unchanged.
The promoted skills are stage-specific and operational.
For example, \textsc{Target-Action Grounding} binds planned actions to source-supported targets and required interface states, while \textsc{Live-DOM Selector Grounding} resolves actions against the rendered page instead of assuming brittle selectors.
During judgment, \textsc{Step-Aligned Evidence} associates observations with the corresponding contract steps before \textsc{Decisive-Evidence Adjudication} determines the verdict.
The supplement lists representative skills and routing relations.

\textbf{Validation-Based Promotion.}\quad
Repairing the residuals that induced a skill does not guarantee that it generalizes. Each candidate is therefore first screened on its local training cases and then evaluated on a disjoint eval set $\mathcal{D}_{\mathrm{eval}}$. It is admitted only when
\begin{equation}
\begin{aligned}
f_{\mathrm{promote}}(k\mid V,\mathcal{D}_{\mathrm{eval}})
={}&[\mathrm{obj}(V\oplus k)>\mathrm{obj}(V)]\\
&\cdot[\mathrm{regress}(V\oplus k)=0]=1,
\end{aligned}
\label{eq:promote}
\end{equation}
where $\mathrm{obj}(\cdot)$ combines fault recall, clean-page specificity, and evaluator-caused inconclusive rate on $\mathcal{D}_{\mathrm{eval}}$, and the no-regression constraint protects cases already handled correctly.
Only validation gains admit a skill.

\textbf{SkillGraph and Routing.}\quad
Promoted skills accumulate into a directed graph $\mathcal{K}$ whose edges encode prerequisites, compatible compositions, and conflicts. Given a requirement $q$ and website $x$, a router activates only the relevant subgraph according to
\begin{equation}
V(q,x) \,=\, V_0 \,\oplus\, \rho(q,x,\, \mathcal{K}), \quad \rho(q,x, \mathcal{K}) \subseteq \mathcal{K}.
\label{eq:route}
\end{equation}
At each stage, the router matches the current Flow Contract and execution artifacts against explicit skill trigger and avoidance predicates, closes prerequisite dependencies, removes incompatible activations through conflict edges, and composes the remaining skills into the stage-specific verifier.
Evolution proceeds through three staged rounds targeting complementary failure sources. Round 1 on $\{z_{\mathrm{plan}},z_{\mathrm{ground}}\}$ improves Flow Contract construction, dependency-independent setup, and target-action grounding. Round 2 on $\{z_{\mathrm{ground}},z_{\mathrm{evid}}\}$ improves executable fixtures, live-DOM interaction, and evidence coverage. Round 3 on $\{z_{\mathrm{judge}}\}$ improves evidence-grounded decisions for persistent and cross-view outcomes. The graph and router are then frozen. Algorithm~\ref{alg:evolution} summarizes evolution.

\begin{algorithm}[t]
\caption{Residual-driven grader evolution.}
\label{alg:evolution}
\textbf{Input}: Base verifier $V_0$, splits $\mathcal{D}_{\mathrm{train}}$, $\mathcal{D}_{\mathrm{eval}}$, and per-round cause subsets $\mathcal{Z}^{(1)}, \mathcal{Z}^{(2)}, \mathcal{Z}^{(3)} \subseteq \mathcal{Z}$.\\
\textbf{Output}: Frozen SkillGraph $\mathcal{K}$ and router $\rho$.
\begin{algorithmic}[1]
\STATE $\mathcal{R}_0\gets f_{\mathrm{run}}(V_0,\mathcal{D}_{\mathrm{train}})$
\STATE $\mathcal{K}\gets f_{\mathrm{induce}}(\mathcal{R}_0),\quad V\gets V_0\oplus\mathcal{K}$
\FOR{$s=1$ to $3$}
\STATE Run the current verifier $V$ on $\mathcal{D}_{\mathrm{train}}$ and collect residuals $\mathcal{R}_s$
\STATE $\mathcal{R}_s^{(s)}\gets\{r\in\mathcal{R}_s\mid f_{\mathrm{attr}}(r)\in\mathcal{Z}^{(s)}\}$
\STATE $\mathcal{K}_s^{\mathrm{cand}}\gets f_{\mathrm{mutate}}(\mathcal{R}_s^{(s)},V)$
\FOR{$k\in\mathcal{K}_s^{\mathrm{cand}}$}
\IF{$f_{\mathrm{screen}}(k,V)=1$ \textbf{and} $f_{\mathrm{promote}}(k\mid V,\mathcal{D}_{\mathrm{eval}})=1$}
\STATE $\mathcal{K}\gets f_{\mathrm{insert}}(\mathcal{K},k,f_{\mathrm{edge}}(k,\mathcal{K}))$
\STATE $V\gets V\oplus k$
\ENDIF
\ENDFOR
\ENDFOR
\STATE $\rho\gets f_{\mathrm{route}}(\mathcal{K})$
\STATE \textbf{return} $\mathcal{K},\ \rho$
\end{algorithmic}
\end{algorithm}

\subsection{Reinforcement Learning with Frozen WebGrader}

After offline evolution, the programmatic verification pipeline, SkillGraph $\mathcal{K}$, and router $\rho$ are frozen as WebGrader. Given a requirement $q$ and policy-generated website $x$, WebGrader first derives $\mathcal{F}_{\mathrm{req}}(q)$, then routes the relevant skill subgraph, grounds every required contract against $x$, and executes it in a live browser. A required flow is marked \textsc{Fail} when it is absent, cannot be grounded or executed, or violates its postconditions. \textsc{Inconclusive} is reserved for evaluator-side ambiguity, exceptions, or malformed verifier output. These evidence-grounded verdicts form the functional component of the RL reward.

\textbf{Programmatic Functional Reward.}\quad
Let the determinate flow set be
\begin{equation}
\mathcal{D}(q,x)=
\{f\in\mathcal{F}_{\mathrm{req}}(q)\mid v_f\in\{\textsc{Pass},\textsc{Fail}\}\}.
\label{eq:determinate}
\end{equation}
WebGrader computes the weighted interaction score
\begin{equation}
I(q,x)=
\frac{\sum_{f\in\mathcal{D}(q,x)}
w_f\,\mathbf{1}[v_f=\textsc{Pass}]}
{\sum_{f\in\mathcal{D}(q,x)}w_f}.
\label{eq:interaction}
\end{equation}
Missing or unexecutable functionality remains in $\mathcal{D}(q,x)$ as a determinate failure and contributes zero to the numerator. Evaluator- or infrastructure-caused inconclusive flows do not enter the functional average. Build or render failures make all required flows fail and yield $I(q,x)=0$.

To encourage both functionality and visual usability, we combine $I(q,x)$ with a frozen appearance score $A(x)\in[0,5]$ as follows.
\begin{equation}
R(q,x)=0.7\!\cdot\!5I(q,x)+0.3A(x).
\label{eq:reward}
\end{equation}
Both components are placed on the same $[0,5]$ scale. The 70/30 mixture prioritizes functional correctness while retaining appearance as a secondary optimization signal.
Across the 12,124 flow verdicts produced during training, 3.09\% remain evaluator-side \textsc{Inconclusive} after website-caused target non-reach is counted as \textsc{Fail}.
The supplement reports the verdict audit and sensitivity results.

\textbf{Group-Relative Optimization.}\quad
For each requirement $q$, the old policy $\pi_{\theta_{\mathrm{old}}}$ samples $G$ websites $\{x_1,\dots,x_G\}$. Frozen WebGrader assigns rewards $R_g=R(q,x_g)$ and computes the group-relative advantage
\begin{equation}
\widehat{A}_g=
\frac{R_g-\operatorname{mean}(R_1,\ldots,R_G)}
{\operatorname{std}(R_1,\ldots,R_G)+\epsilon}.
\label{eq:grpo-advantage}
\end{equation}
GRPO~\cite{shao2024deepseekmath} updates $\pi_\theta$ through the standard clipped surrogate objective with a KL penalty toward a reference policy. With WebGrader frozen, rollout rewards differ only through executable outcomes and require no critic.

\section{Experiments}

\subsection{Benchmarks and Metrics}

\textbf{WebGen-Bench.}\quad The public test set contains 101 website requirements and 647 manually curated functional cases \cite{lu2025webgenbench}. We report valid render rate (VRR), average appearance score (AAS, from 1 to 5), and functional success rate with 0.5 credit for \textsc{Partial} outcomes
\begin{equation}
\mathrm{FSR}=\frac{N_{\mathrm{yes}}+0.5\,N_{\mathrm{partial}}}{647}.
\label{eq:fsr}
\end{equation}

\textbf{WG-core-250.}\quad We further use a fixed 250-task subset of HTMLBench-400 \cite{wu2026htmlcure} and report its Full Score, which combines rendering, GPT-5.4 visual quality, functionality, interactivity, and code quality, together with deterministic browser test-case pass rate (TC\%). Thus WebGen-Bench and WG-core-250 probe the policy with a UI agent and deterministic browser cases, respectively. Composition and scoring details are in the supplement.

WebGen-Verifier-100 has ten construction categories versus six overlapping WebGen-Bench categories.

\textbf{Verifier Analysis.}\quad We analyze the frozen verifier on the WebGen-Verifier-100 Test split defined in Method.
Detailed construction, per-stage metrics, and qualification protocols are provided in the supplement.

\subsection{Experimental Setup}

We initialize WebGrader-RL from a Qwen3-8B checkpoint obtained after one epoch of full-parameter SFT on 600 rebuilt silver examples~\cite{yang2025qwen3}.
The SFT queries are stratified samples from the public WebGen-Instruct train split, and Claude Sonnet 4.6 supplies browser-executable React/Vite targets in the WebGen-R1 artifact format.
For RL, we use a different set of 600 queries independently sampled from the same training split and optimize for 100 effective VERL/GRPO updates~\cite{lu2025webgenbench,sheng2024hybridflow,shao2024deepseekmath}.
All 101 WebGen-Bench test queries remain held out, and each update samples four RL queries with eight completions each.
DeepSeek-v4-flash implements the four-stage verifier and GPT-5.4-mini supplies the appearance reward, both through public APIs.
Controlled baselines share the initialization, data, rollout budget, sampling, and optimization while differing only in reward construction.
GUI-RL uses model-based interaction, VLM+GUI-RL adds appearance, and VLM+Base-Script-RL uses the unevolved script, isolating verifier evolution.
Every reported generator is re-evaluated from scratch under the same frozen benchmark pipelines.
For WebGen-Bench, the GPT-5.4 evaluator version, prompts, configuration, and browser settings are fixed across systems, with content hashes retained.

\subsection{Main Results}

\begin{table}[t]
\centering
{\small
\setlength{\tabcolsep}{2.5pt}
\begin{tabular}{lccc@{\hspace{4pt}}cc}
\toprule
& \multicolumn{3}{c}{WebGen-Bench} & \multicolumn{2}{c}{WG-core-250}\\
\cmidrule(lr){2-4}\cmidrule(lr){5-6}
Method & VRR $\uparrow$ & AAS $\uparrow$ & FSR $\uparrow$ & Score $\uparrow$ & TC\% $\uparrow$\\
\midrule
\multicolumn{6}{l}{\emph{External generator context}}\\
o4-mini & 89.11 & 2.49 & 40.65 & 39.068 & 31.668\\
DeepSeek-v4-flash & 93.07 & 2.95 & 47.91 & 43.684 & 36.103\\
Qwen3-Coder-480B & 94.06 & 3.32 & 53.25 & 42.504 & 36.328\\
GPT-5.4-mini & 89.11 & 3.27 & 52.32 & 49.636 & 42.277\\
Claude Opus 4.8 & 100.00 & 3.96 & 68.20 & 50.460 & 45.653\\
GPT-5.5 & 90.10 & 3.78 & 62.44 & 52.280 & 47.783\\
\midrule
\multicolumn{6}{l}{\emph{Controlled training comparison}}\\
Qwen3-8B base & 90.10 & 1.51 & 14.92 & 31.160 & 27.418\\
WebGen SFT & 62.38 & 2.19 & 33.62 & 32.388 & 25.640\\
GUI-RL & 96.04 & 3.15 & 41.96 & 38.120 & 33.861\\
VLM+GUI-RL & 100.00 & 3.43 & 42.66 & 37.746 & 29.345\\
VLM+Base-Script-RL & 100.00 & 3.46 & 44.13 & 38.372 & 33.525\\
\textbf{WebGrader-RL (8B)} & \textbf{100.00} & \textbf{3.48} & \textbf{52.01} & \textbf{44.953} & \textbf{39.931}\\
\bottomrule
\end{tabular}}
\caption{Main results on WebGen-Bench and WG-core-250. \textbf{Bold} marks the best controlled result.}
\label{tab:main-results}
\end{table}

Relative to the matched VLM+Base-Script-RL baseline, WebGrader-RL improves FSR by 7.88 points while changing AAS by only 0.02, indicating that the gain is primarily functional. On WG-core-250 it adds 6.58 Full Score and 6.41 TC points over the same baseline, supporting transfer to an independently designed evaluator. The 8B policy also exceeds o4-mini and DeepSeek-v4-flash under both protocols, while remaining behind the strongest closed-source generators.

\section{Analysis}

Residual-driven evolution discovers localized verifier repairs, routing composes them where needed, and broader browser coverage produces more reliable functional rewards.
The following analyses connect verifier structure and execution behavior to downstream transfer.

\subsection{Routed Evolution Produces Reusable Repairs}

A flat-rule control separates structured evolution from simply adding more instructions.
On the same WebGen-Verifier-100 Test cases, we compare the base verifier $V_0$, a flat prompt containing every discovered rule, the routed SkillGraph, and the graph with conflict handling.
As shown in Table~\ref{tab:budget}, the flat rule set increases token usage from 99.5K to 128.5K but improves macro-F1 only from 0.7813 to 0.7896.
Under a similar token budget, routing raises F1 to 0.9037.
Conflict handling further increases it to 0.9248 while reducing inconclusive outcomes to 4\%.
On the matched-budget cases, conflict handling retains 1.25 skills per example instead of applying all four candidates.
Across the broader routing audit, the router activates an average of 1.46 skills with 84.2\% activation precision and 88.6\% recall.
Macro-F1 averages fault- and clean-class F1, counting an evaluator-side inconclusive verdict as an error for its ground-truth class.
These results suggest that selecting and composing relevant skills contributes more than prompt length alone.

\begin{table}[t]
\centering
{\small
\setlength{\tabcolsep}{1.8pt}
\begin{tabular}{@{}lccccc@{}}
\toprule
Structure & F1 & Recall & Specificity & Inconclusive & Tokens\\
\midrule
$V_0$ & 0.7813 & 0.76 & 0.80 & 0.18 & 99.5K\\
Flat rules & 0.7896 & 0.76 & 0.80 & 0.16 & 128.5K\\
SkillGraph & 0.9037 & 0.88 & 0.92 & 0.06 & 126.1K\\
Graph + conflicts & \textbf{0.9248} & \textbf{0.92} & \textbf{0.92} & \textbf{0.04} & 126.7K\\
\bottomrule
\end{tabular}}
\caption{Verifier structures on the WebGen-Verifier-100 Test split. Token counts are totals over the same cases, and Test outcomes are used only for reporting.}
\label{tab:budget}
\end{table}

The improvement also accumulates across the three evolution rounds.
Planning and grounding first raise fault recall from 0.650 to 0.744 and recover 15 additional faults.
Executable fixtures and evidence collection raise recall to 0.800 and recover another nine faults.
Evidence-grounded judgment produces the final increase to 0.850 and recovers eight additional faults.
Clean-page specificity rises simultaneously from 0.80 to 0.95.
Together with validation-based promotion, these localized repairs progressively improve verifier quality before the SkillGraph is frozen.

\begin{table}[t]
\centering
{\small
\setlength{\tabcolsep}{2.8pt}
\begin{tabular}{lrrrr}
\toprule
Stage & Hits & Recall & Fault target cov. & Clean spec.\\
\midrule
$V_0$ & 104 & 0.650 & 0.635 & 0.80\\
Planning/grounding & 119 & 0.744 & 0.748 & 0.85\\
Grounding/evidence & 128 & 0.800 & 0.817 & 0.90\\
Judgment & 136 & 0.850 & 0.878 & 0.95\\
\bottomrule
\end{tabular}}
\caption{Cumulative Test performance across evolution. Fault target coverage is measured over injected fault variants at each saved stage.}
\label{tab:stagewise}
\end{table}

Two residual-derived patterns make these repairs operational.
\textsc{Executable Fixture} creates missing prerequisites or persistent state, while \textsc{Step-Aligned Evidence} binds DOM, response, or persistent-state observations to the responsible step without page-specific selectors.
They also transfer beyond their source pages.
The former activates 121 times across 18 pages and repairs 13 cases. The latter spans seven categories and eight flow families, repairs 11 cases, and yields the largest local lift of 0.128.

Leave-one-skill-out interventions test whether the routed modules materially affect verdicts.
Disabling a measured skill changes 8 to 14 Test verdicts, compared with two changes under an unchanged repeat control.
All six measured skills have positive net correctness, and five interventions exceed repeat drift.
Evidence checkpoints produce the largest net contribution (+7), followed by target-action grounding and executable fixtures (+6 each).
This ties the graph's gain to operational repairs.

\subsection{Evolution Improves Execution Before Judgment}

The Test split shows where these routed repairs change verification.
On the WebGen-Verifier-100 Test split, WebGrader detects 136 of 160 faults and accepts 19 of 20 clean pages.
Compared with $V_0$, it raises fault recall from 65\% to 85\% and clean-page specificity from 80\% to 95\%.
It also reduces evaluator-caused inconclusive outcomes from 13.3\% to 3.9\%, resulting in 99.3\% precision.
WebGrader consequently provides a substantially more accurate verification signal than the original script, rule grader, and GUI grader.

\begin{table}[t]
\centering
{\small
\setlength{\tabcolsep}{3.0pt}
\begin{tabular}{lcccc}
\toprule
Verifier & Specificity & Recall & Precision & Inconclusive\\
\midrule
Rule grader & 90.0 & 20.0 & 94.1 & 5.0\\
GUI grader & 80.0 & 60.0 & 96.0 & 15.0\\
$V_0$ (base script) & 80.0 & 65.0 & 96.3 & 13.3\\
\textbf{WebGrader} & \textbf{95.0} & \textbf{85.0} & \textbf{99.3} & \textbf{3.9}\\
\bottomrule
\end{tabular}}
\caption{Verifier performance (\%) on the WebGen-Verifier-100 Test split.}
\label{tab:verifier}
\end{table}

Stage-level diagnostics locate this improvement before the final verdict.
The largest gains occur in strong-evidence collection (+15.6), mandatory-observation coverage (+14.9), post-action state reach (+14.4), and target-action execution (+13.6).
The stage-level gains show that WebGrader improves verification by reaching the decisive state and observing its consequences before semantic judgment.
Executor success also rises by 8.9 points, consistent with improved evidence being paired with more reliable action grounding.

\begin{table}[t]
\centering
{\small
\setlength{\tabcolsep}{4pt}
\begin{tabular}{lccc}
\toprule
Execution diagnostic & Original & Evolved & $\Delta$\\
\midrule
Executor success & 90.0 & \textbf{98.9} & +8.9\\
Post-action state reached & 74.2 & \textbf{88.6} & +14.4\\
Target action executed & 78.1 & \textbf{91.7} & +13.6\\
Strong evidence collected & 71.6 & \textbf{87.2} & +15.6\\
Must-collect coverage & 75.4 & \textbf{90.3} & +14.9\\
\bottomrule
\end{tabular}}
\caption{Execution diagnostics (\%) on Test. The denominator comprises planned requirement-level flows. Target action means successful dispatch. Post-action state additionally requires the intended result to be observed.}
\label{tab:diagnostics}
\end{table}

Gains increase with interaction complexity.
Recall improves by 20.0 points over all faults, by 24.3 points on multi-action stateful faults, and by 28.1 points when this subset is restricted to medium- and high-severity cases.
These are the cases in which correctness depends most strongly on prerequisite setup, target-action execution, and postcondition inspection.
Further decompositions appear in the supplement.

Slice-level results sharpen this picture.
Recall rises by 20.6 points on latent faults, 21.3 on stateful faults, and 25.0 on depth-2-to-3 flows.
The remaining errors concentrate in depth-4+ flows (0.789 recall), data binding (0.789), and search/filter/sort (0.810).
WebGrader repairs failures requiring hidden state changes, while long trajectories and diffuse data dependencies remain hardest.

\subsection{Better Verification Transfers to Policy Learning}

The stronger verifier also produces broader policy gains.
Table~\ref{tab:main-results} shows that WebGrader-RL improves WebGen-Bench FSR by 7.88 points over the matched VLM+Base-Script-RL baseline while changing AAS by only 0.02, indicating that the gain is primarily functional.
Table~\ref{tab:category-attribution} further shows gains across all six categories, ranging from 4.73 points on data management to 15.20 on design validation.
The evolved reward therefore improves a broad range of website behaviors.
Because category labels are not used for routing or reward aggregation, this spread also indicates that reusable execution skills transfer across task semantics.

\begin{table}[t]
\centering
{\small
\setlength{\tabcolsep}{3.8pt}
\begin{tabular}{lccc}
\toprule
Category & Base script & WebGrader & $\Delta$\\
\midrule
Content presentation & 54.31 & \textbf{63.80} & +9.49\\
User interaction & 41.69 & \textbf{48.20} & +6.51\\
Data management & 41.88 & \textbf{46.61} & +4.73\\
Functional testing & 30.68 & \textbf{36.00} & +5.32\\
Data display & 55.11 & \textbf{61.60} & +6.49\\
Design validation & 66.80 & \textbf{82.00} & +15.20\\
\bottomrule
\end{tabular}}
\caption{Category-wise FSR (\%) on WebGen-Bench for matched VLM+Base-Script-RL and WebGrader-RL.}
\label{tab:category-attribution}
\end{table}

The improvement persists under an independently designed evaluation protocol.
On WG-core-250, WebGrader-RL raises deterministic browser test pass rate by 6.41 points and aggregate score by 6.58 over the matched base-script reward.
The gains transfer beyond WebGrader's evaluator.

Paired analyses confirm that this transfer is not driven by a small number of tasks.
On WebGen-Bench, query-macro FSR improves by 5.7 points with a 95\% paired-bootstrap interval of $[+1.9,+9.5]$ and permutation $p=.004$.
On WG-core-250, task-macro Full Score and test-case ratio improve by 11.2 and 8.8 points, their intervals exclude zero, and the exact McNemar test gives $p=7.3\times10^{-5}$.
The gains are therefore consistent across paired queries and tasks as well as across evaluator designs.

\section{Conclusion}

WebGrader turns open-ended requirements into functional rewards grounded in observed browser state transitions.
Its requirement-first verifier derives the necessary flows, grounds them against each generated website, and judges post-action evidence.
A residual-driven loop promotes localized repairs into a routed SkillGraph that is fixed before policy training.
Across UI-agent and deterministic browser evaluations, WebGrader-RL improves an 8B policy beyond matched script-based supervision.
Analysis attributes these gains to better prerequisite setup, target-action execution, and post-action evidence collection while preserving clean-page specificity.
These results suggest a broader principle for agentic RL.
When task success is revealed through interaction, reward construction should make the required trajectory and its observable consequences explicit before optimization.
Together, executable reward construction connects open-ended generation to policy optimization, translating verifier improvements into more reliable functional behavior.

\clearpage
\bibliography{references}

\clearpage
\section*{Supplementary Material}
\def\webgraderappendix{1}
\ifdefined\webgraderappendix\else
\documentclass[letterpaper]{article} 
\usepackage{aaai2027}
\usepackage[hyphens]{url} 
\usepackage{graphicx} 
\urlstyle{rm} 
\def\UrlFont{\rm} 
\usepackage{natbib} 
\usepackage{caption} 
\usepackage{booktabs}
\usepackage{amsmath}
\frenchspacing 

\pdfinfo{
/TemplateVersion (2027.1)
}
\nocopyright

\setcounter{secnumdepth}{0}

\title{WebGrader: Supplementary Material}
\author{
    Boshui Chen\textsuperscript{\rm 1},
    Huiping Liu\textsuperscript{\rm 2},
    Shaolei Zhang\textsuperscript{\rm 3}\corresponding
}
\affiliations{
    \textsuperscript{\rm 1}Beijing Institute of Technology, China\\
    \textsuperscript{\rm 2}Beijing Jiaotong University, China\\
    \textsuperscript{\rm 3}Renmin University of China, China\\
    zhangshaolei98@ruc.edu.cn
}

\begin{document}
\maketitle
\fi

\section{Reproducible Verifier Specification}

\subsection{Artifacts and Stage Interfaces}

WebGrader first derives required interaction flows from the requirement alone.
For each required flow, it then materializes a Flow Contract and scores the website from its execution evidence.
The contract exposes the intended preconditions, action sequence, target action, evidence checkpoints, postconditions, mandatory observations, and criticality weight.
The planner produces these contracts.
The grounding stage compiles them into Playwright programs against the generated project's source and live DOM.
The evidence stage executes the programs and records structured browser traces.
The frozen evidence-conditioned judge then returns one semantic verdict per flow.

\begin{center}
\centering
{\small
\setlength{\tabcolsep}{3.2pt}
\begin{tabular}{p{0.27\columnwidth}p{0.62\columnwidth}}
\toprule
Artifact & Core contents\\
\midrule
Flow Contract & Preconditions, fixtures, action sequence, target action, evidence checkpoints, postconditions, mandatory observations, and criticality weight\\
Executable script & Grounded Playwright selectors and actions, state-reset operations, and evidence-capture hooks\\
Evidence bundle & Screenshots, DOM snapshots, responses, URL changes, console/network events, storage state, refresh outcomes, and assertion traces\\
Flow verdict & \textsc{Pass}, \textsc{Fail}, or \textsc{Inconclusive}, with evidence pointers and a stage-specific residual when applicable\\
\bottomrule
\end{tabular}}
\captionof{table}{Verifier artifacts passed across the four stages.}
\label{tab:supp-verifier-artifacts}
\end{center}

\subsection{Runtime and Verdict Contract}

The browser harness records state reset, navigation, action attempts and resolution strategy, target-action reach, screenshots, DOM state, response messages, URL changes, console and network events, local/session storage, refresh outcomes, and final assertions.
The judge receives the requirement, Flow Contract, and execution trace, but never the benchmark's private clean/fault label.

\begin{center}
\centering
{\small
\setlength{\tabcolsep}{3.8pt}
\begin{tabular}{lr}
\toprule
Operation & Timeout\\
\midrule
Render navigation & 15 s, network idle\\
Interactive navigation & 30 s, DOM content loaded\\
Playwright action & 7 s\\
Single flow & 20 s\\
Flow process & 25 s\\
Completion/case & 120 s\\
Verifier API main/fallback & 60 s / 15 s\\
Appearance VLM main/fallback & 10 s / 10 s\\
Appearance VLM total & 22 s\\
\bottomrule
\end{tabular}}
\captionof{table}{Frozen runtime timeouts for distinguishing site failures from evaluator uncertainty and bounding per-completion cost.}
\label{tab:supp-timeouts}
\end{center}

\textsc{Pass} requires successful target-action execution and sufficient evidence that the requested postcondition holds.
\textsc{Fail} covers executed evidence of a semantic violation and model-caused inability to build, render, expose a required control, reach the required flow, or execute the target action.
A missing control or unexecutable target action is therefore a determinate website failure.
\textsc{Inconclusive} is reserved for evaluator-side uncertainty, including unresolved selector ambiguity, browser or infrastructure exceptions, timeouts not attributable to the generated website, and malformed verifier output.

\section{Representative Skills and Routing}

The concrete stage-specific registries contain 6 planning skills, 8 execution/evidence skills, and 7 judgment skills.
We show representative entries that expose their activation signals, localized operations, guards, and routing relations.

\begin{center}
\centering
{\small
\setlength{\tabcolsep}{2pt}
\begin{tabular}{p{0.10\columnwidth}p{0.23\columnwidth}p{0.24\columnwidth}p{0.32\columnwidth}}
\toprule
Stage & Skill & Activation signal & Local operation and guard\\
\midrule
Plan & \textsc{Coverage Inventory} & Multiple functional obligations & Separates independently checkable flows without repeatedly testing the same target\\
Plan & \textsc{Target-Action Grounding} & Target must be recovered from source or UI state & Binds the action to a source-supported target and prerequisite state without inventing a control or selector\\
Exec. & \textsc{Live-DOM Selector Grounding} & A preset selector is absent, unstable, or ambiguous & Enumerates the live DOM and grounds by semantics, role, and attributes without converting evaluator ambiguity into site failure\\
Exec. & \textsc{Flow-Isolated Reset} & Flows may share mutable state & Restores page and persistent state between flows while preserving required fixtures\\
Judge & \textsc{Step-Aligned Evidence} & Evidence spans multiple actions or snapshots & Associates each observation with the contract step that produced it to prevent cross-step attribution\\
Judge & \textsc{Decisive-Evidence Adjudication} & Evidence sources differ in diagnostic strength & Uses the most discriminative direct evidence so weak visual cues cannot override explicit state evidence\\
Judge & \textsc{Evidence-Sufficiency Check} & Neither pass nor fail has decisive support & Preserves \textsc{Inconclusive} when evaluator evidence is missing\\
\bottomrule
\end{tabular}}
\captionof{table}{Representative promoted skills. Names summarize the operational behavior, and internal registry identifiers are omitted.}
\label{tab:supp-skill-examples}
\end{center}

Routing follows the executable dependency chain
{\footnotesize
\[
\begin{aligned}
\textsc{Coverage Inventory}
&\rightarrow \textsc{Target-Action Grounding}\\
&\rightarrow \textsc{Live-DOM Selector Grounding}\\
&\rightarrow \textsc{Step-Aligned Evidence}\\
&\rightarrow \textsc{Decisive-Evidence Adjudication}.
\end{aligned}
\]
}
\textsc{Flow-Isolated Reset} composes with execution skills when flows share state.
\textsc{Evidence-Sufficiency Check} conflicts with an unsupported determinate verdict and takes precedence when the decisive observation is missing because of evaluator-side uncertainty.
The complete registries additionally encode prerequisites, compatible compositions, conflict edges, negative constraints, and validation-promotion provenance.

\section{WebGen-Verifier-100 Protocol}

\subsection{Benchmark Construction}

WebGen-Verifier-100 is constructed from 100 public WebGen-Instruct training queries stratified across ten application categories. For each query, a React/Vite implementation is accepted only when it can be installed, built, served, and rendered locally. Its core requirement-level flows must expose reachable target actions and observable outcomes, and it must contain no blocking runtime error. Eight source-level single-fault variants are then materialized from naturally occurring WebDev failure patterns. The normalized families cover validation and authentication, search/filter/sort, state persistence, cross-view synchronization, binding and routing, calculation and aggregation, CRUD workflows, and feedback visibility. A variant is active only after matched execution of the clean and faulty versions confirms that its target flow remains reachable and the intended behavioral difference is observable, isolated, and free of explicit markers. Twenty-two ambiguous, unstable, or out-of-scope variants are excluded or quarantined, leaving 778 active faults.

The split is performed at the page level. A clean page, all its faults, its Flow Contracts, and its private references remain in one split. The ten application categories are analytics/dashboard, CRM, e-commerce, ERP, job search, learning, productivity, project management, social media, and travel. Each category contributes ten pages, split 6/2/2 across Train/Eval/Test.

\begin{center}
\centering
{\small
\setlength{\tabcolsep}{7pt}
\begin{tabular}{lrrrr}
\toprule
Item & Train & Eval & Test & Total\\
\midrule
Clean pages & 60 & 20 & 20 & 100\\
Active single-fault pages & 461 & 157 & 160 & 778\\
Active flows & 282 & 95 & 93 & 470\\
Easy construction & 172 & 59 & 60 & 291\\
Medium construction & 173 & 59 & 60 & 292\\
Hard construction & 116 & 39 & 40 & 195\\
\bottomrule
\end{tabular}}
\captionof{table}{Page-level composition of WebGen-Verifier-100. Easy, medium, and hard denote construction difficulty. The Test benchmark contains 20 clean and 160 active single-fault pages.}
\label{tab:supp-composition}
\end{center}

Train60 is used to induce residuals and construct candidate mutations, while Eval20 serves as the disjoint validation set for candidate promotion.
Test20 is reserved exclusively for reporting verifier performance.
Its private labels, fault specifications, expected transitions, and reference evidence remain inaccessible to the verifier and optimization process.
Test outcomes are never used for skill induction, mutation, promotion, routing design, early stopping, or hyperparameter selection.
Reported intermediate and final verifier measurements on Test20 are evaluation-only checkpoints and are never fed back into development.

Ground truth has five layers comprising page metadata and the natural-language requirement, Flow Contracts, isolated fault specifications, reference action/evidence traces, and scoring metadata. A reference trace defines an execution-semantic contract without prescribing a unique Playwright program. Alternative action sequences are valid when they establish the required preconditions, execute the target action, cover the mandatory evidence, and expose the same clean invariant or fault differential. Public artifacts contain only the requirement and generated project. Private artifacts contain the clean/fault label, flow identifiers, target-action obligations, and reference evidence. This separation prevents the deployed verifier from reading its answer while preserving a complete audit trail for offline evolution and evaluation.

Construction difficulty describes how difficult a controlled fault is to materialize and validate. Separately, behavioral severity describes its effect on the requested behavior. The Test split contains 29 low-, 76 medium-, and 55 high-severity faults. Of the 470 active flows, 438 have criticality weight 2 and 32 have weight 3. All Test flows have weight 2. A clean application is accepted only when its core flows execute without a critical semantic failure. A fault variant is retained only when the corresponding target behavior differs from the clean reference.

\subsection{Qualitative Execution Trace}

\begin{figure*}[t]
\centering
\includegraphics[width=0.98\textwidth]{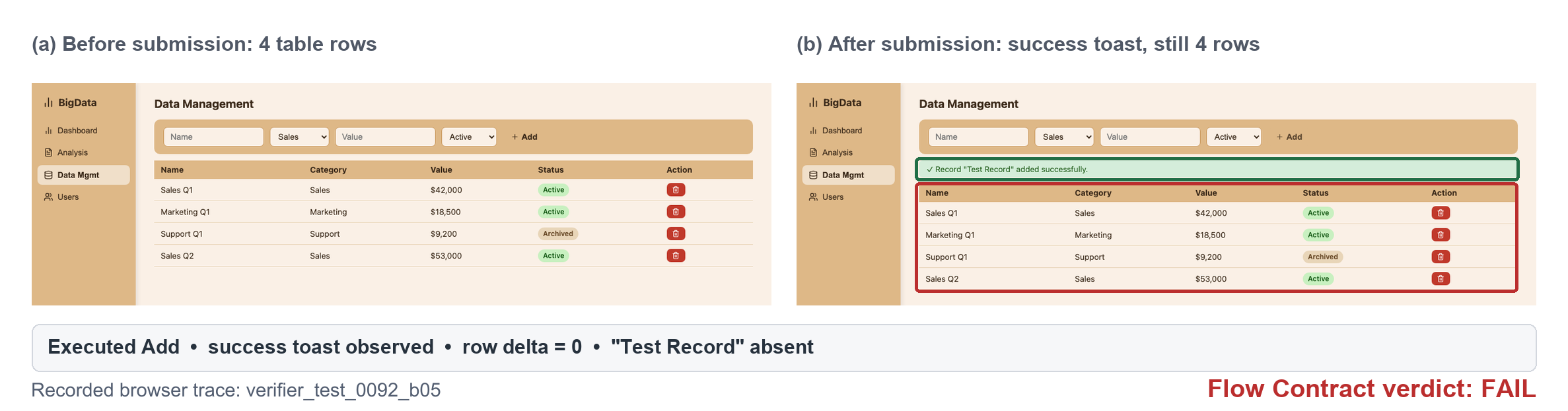}
\caption{Frozen replay for fault \texttt{verifier\_test\_0092\_b05}. This neutral public identifier replaces the internal construction identifier. The target action completes and the page displays a success toast, but the required table update does not occur.}
\label{fig:supp-trace-case}
\end{figure*}

The case in Figure~\ref{fig:supp-trace-case} is taken directly from the benchmark's real browser trace. Its action and evidence record appears below.
\begin{center}
\centering
{\small
\setlength{\tabcolsep}{3pt}
\begin{tabular}{p{0.34\columnwidth}p{0.55\columnwidth}}
\toprule
Evidence field & Recorded value\\
\midrule
Target action done & true\\
Rows before / after & 4 / 4\\
Success response & Record ``Test Record'' added successfully\\
Requested record in table & false\\
Row delta & 0\\
\bottomrule
\end{tabular}}
\captionof{table}{Evidence supporting the \textsc{Fail} verdict in the qualitative trace.}
\label{tab:supp-trace-evidence}
\end{center}

\section{Evolution Mechanism and Diagnostics}

\subsection{Candidate Mutation and Promotion}

The initial verifier $V_0$ uses DeepSeek-v4-flash for the four-stage pipeline. An offline ground-truth-aware Claude Opus 4.8 meta-evaluator attributes residuals, clusters recurring failures, and proposes skill mutations. It can access private annotations only during offline verifier evolution and is not part of blind verification or the RL reward. The base model weights are unchanged. Candidate skills modify structured verifier artifacts covering planner evidence obligations, target-action constraints, fixtures, DOM-grounding strategies, and semantic judgment rules.

\begin{center}
\centering
{\small
\setlength{\tabcolsep}{5pt}
\begin{tabular}{lrr}
\toprule
Candidate & Flow F1 & Recall\\
\midrule
R03 & .9214 & .9182\\
M01 & .9278 & .9235\\
M02 & .9341 & .9306\\
M03 & \textbf{.9432} & .9378\\
M04 & .9397 & \textbf{.9415}\\
M05 & .9316 & .9289\\
\bottomrule
\end{tabular}
\par\smallskip
\setlength{\tabcolsep}{4pt}
\begin{tabular}{lrrr}
\toprule
Candidate & Evidence & Grounding & Over-gen.\\
\midrule
R03 & .8675 & .9268 & .0251\\
M01 & .8742 & .9310 & .0228\\
M02 & .8819 & .9364 & .0207\\
M03 & .8894 & .9412 & .0193\\
M04 & \textbf{.9018} & \textbf{.9481} & \textbf{.0146}\\
M05 & .8837 & .9345 & .0215\\
\bottomrule
\end{tabular}}
\captionof{table}{Candidate mutation screening on Train60 for the final promotion step. M04 trades a small amount of Flow F1 for the best recall, evidence, grounding, and over-generation profile.}
\label{tab:supp-train-candidates}
\end{center}

\begin{center}
\centering
{\small
\setlength{\tabcolsep}{3.2pt}
\begin{tabular}{lrrrrr}
\toprule
Eval variant & F1 & Rec. & Evid. & Ground. & Over.\\
\midrule
Base & .8425 & .8516 & .7784 & .9012 & .0385\\
Base-skill & .8917 & .8978 & .8426 & .9285 & .0241\\
M03 & .9286 & .9164 & .8861 & .9428 & .0197\\
M04 / evolved & \textbf{.9479} & \textbf{.9442} & \textbf{.9085} & \textbf{.9516} & \textbf{.0138}\\
\bottomrule
\end{tabular}}
\captionof{table}{Validation-set promotion on Eval20. M04 is selected without using Test20 outcomes.}
\label{tab:supp-eval-promotion}
\end{center}

\subsection{Stage-Wise Accumulation}

\begin{center}
\centering
{\small
\setlength{\tabcolsep}{4pt}
\begin{tabular}{lrrr}
\toprule
Stage & Hits & Recall & Fault target cov.\\
\midrule
$V_0$ & 104 & .650 & .635\\
Planning/grounding & 119 & .744 & .748\\
Grounding/evidence & 128 & .800 & .817\\
Judgment & 136 & .850 & .878\\
\bottomrule
\end{tabular}
\par\smallskip
\setlength{\tabcolsep}{6pt}
\begin{tabular}{lrrr}
\toprule
Stage & Clean spec. & Clean FPR & AUROC\\
\midrule
$V_0$ & .80 & .15 & .760\\
Planning/grounding & .85 & .10 & .810\\
Grounding/evidence & .90 & .05 & .854\\
Judgment & .95 & .05 & .890\\
\bottomrule
\end{tabular}}
\captionof{table}{Evaluation-only Test20 performance of saved verifier checkpoints across the three evolution rounds. Fault target coverage is measured over injected fault variants at each saved stage. The end-to-end execution diagnostics instead use planned requirement-level flows. Test outcomes are not fed back into development.}
\label{tab:supp-cumulative}
\end{center}

\begin{center}
\centering
{\small
\setlength{\tabcolsep}{2.5pt}
\begin{tabular}{lrrrr}
\toprule
Transition & Recovered & Regressed & Net & Paired acc.\\
\midrule
$V_0\rightarrow$ planning & 18 & 3 & +15 & +.061\\
Planning $\rightarrow$ evidence & 11 & 2 & +9 & +.055\\
Evidence $\rightarrow$ judgment & 10 & 2 & +8 & +.048\\
$V_0\rightarrow$ final & 39 & 7 & +32 & +.164\\
\bottomrule
\end{tabular}}
\captionof{table}{Recovered and regressed Test faults across evolution stages.}
\label{tab:supp-transitions}
\end{center}

\begin{center}
\centering
{\small
\setlength{\tabcolsep}{4pt}
\begin{tabular}{lrrr}
\toprule
Target scope & $n$ & Original & Evolved\\
\midrule
All faults & 160 & .650 & .850\\
Multi-action + stateful & 115 & .635 & .878\\
Multi-action + stateful + med/high & 96 & .625 & .906\\
\bottomrule
\end{tabular}}
\captionof{table}{Recall on progressively more interaction-intensive Test scopes.}
\label{tab:supp-target-scope}
\end{center}

\subsection{Graph Structure at Matched Budget}

\begin{center}
\centering
{\small
\setlength{\tabcolsep}{3pt}
\begin{tabular}{lrrrr}
\toprule
Structure & F1 & Recall & Spec. & Inc.\\
\midrule
$V_0$ & .7813 & .76 & .80 & .18\\
Flat rules & .7896 & .76 & .80 & .16\\
Skill graph & .9037 & .88 & .92 & .06\\
Graph + conflicts & \textbf{.9248} & \textbf{.92} & \textbf{.92} & \textbf{.04}\\
\bottomrule
\end{tabular}
\par\smallskip
\setlength{\tabcolsep}{6pt}
\begin{tabular}{lrr}
\toprule
Structure & Tokens & Overhead\\
\midrule
$V_0$ & 99,480 & 0.0\%\\
Flat rules & 128,530 & 29.2\%\\
Skill graph & 126,140 & 26.8\%\\
Graph + conflicts & 126,720 & 27.4\%\\
\bottomrule
\end{tabular}}
\captionof{table}{Matched-budget analysis on the full WebGen-Verifier-100 Test split. Test outcomes are used only for reporting. A flat prompt spends slightly more tokens than the graph while yielding little improvement over $V_0$. Selective routing and conflict handling produce the gain.}
\label{tab:supp-matched-budget}
\end{center}

Flat rules apply all four candidates to every Test case, for a mean of 4.0 active rules. The graph retrieves the same four candidates but activates 1.42 skills per case on average. Conflict handling lowers this to 1.25. Thus its advantage is conditional composition, not additional inference.

\subsection{Leave-One-Skill-Out Interventions}

\begin{center}
\centering
{\small
\setlength{\tabcolsep}{3.5pt}
\begin{tabular}{lrrrr}
\toprule
Disabled skill & Active & Changes & Nec./harm. & Net\\
\midrule
Evidence checkpoints & 82 & 14 & 9 / 2 & +7\\
Dependency setup & 76 & 11 & 7 / 2 & +5\\
Target-action grounding & 58 & 10 & 7 / 1 & +6\\
Executable fixtures & 91 & 13 & 8 / 2 & +6\\
Decisive evidence & 69 & 8 & 5 / 1 & +4\\
Step-aligned evidence & 61 & 7 & 4 / 1 & +3\\
\bottomrule
\end{tabular}}
\captionof{table}{Leave-one-skill-out interventions on the WebGen-Verifier-100 Test split. ``Necessary'' counts correct evolved verdicts lost when the skill is disabled. ``Harmful'' counts errors repaired by disabling it. The unchanged repeat control changes only two verdicts.}
\label{tab:supp-loo}
\end{center}

The skill-off runs change between 8 and 14 verdicts, compared with two under repeat execution. Five of six interventions exceed this drift. Every measured skill has positive net correctness, with the largest contribution from explicit evidence checkpoints.

\subsection{Transfer and Routing}

\begin{center}
\centering
{\small
\setlength{\tabcolsep}{3.2pt}
\begin{tabular}{lrrrr}
\toprule
Skill & Pages & Active & Cats. & Flows\\
\midrule
Evidence & 12 & 86 & 7 & 8\\
Dependency & 11 & 79 & 7 & 7\\
Target action & 9 & 62 & 6 & 7\\
Fixtures & 18 & 121 & 10 & 8\\
Live DOM & 17 & 108 & 9 & 8\\
Persistent state & 13 & 59 & 8 & 6\\
Decisive evidence & 14 & 73 & 8 & 7\\
Step-aligned evidence & 12 & 64 & 7 & 6\\
\bottomrule
\end{tabular}
\par\smallskip
\setlength{\tabcolsep}{7pt}
\begin{tabular}{lrr}
\toprule
Skill & Local lift & Repairs\\
\midrule
Evidence & +.128 & +11\\
Dependency & +.089 & +7\\
Target action & +.097 & +6\\
Fixtures & +.107 & +13\\
Live DOM & +.065 & +7\\
Persistent state & +.034 & +2\\
Decisive evidence & +.055 & +4\\
Step-aligned evidence & +.047 & +3\\
\bottomrule
\end{tabular}}
\captionof{table}{Skill transfer across pages, application categories, and flow families. Improvements are not confined to the pages that originally induced each skill.}
\label{tab:supp-transfer}
\end{center}

The router has 84.2\% activation precision, 88.6\% recall, a 1.1\% invalid-activation rate, and 1.46 active skills per case. Conflict resolution suppresses 7.2\% of candidate activations, while 6.8\% of retained activations are judged unnecessary. Together with Table~\ref{tab:supp-matched-budget}, these results show that the graph is both selective and broadly reusable.

\subsection{Blind-Spot Decomposition}

\begin{center}
\centering
{\small
\setlength{\tabcolsep}{2.2pt}
\begin{tabular}{lrrr}
\toprule
Slice ($n$) & Orig. & Evol. & Gain\\
\midrule
All (160) & .650 & .850 & +.200\\
Immediately visible (19) & .789 & .947 & +.158\\
Latent (141) & .631 & .837 & +.206\\
Single-action (15) & .800 & .933 & +.133\\
Multi-action (145) & .634 & .841 & +.207\\
Stateless (38) & .737 & .895 & +.158\\
Stateful (122) & .623 & .836 & +.213\\
Depth 1 (15) & .800 & .933 & +.133\\
Depth 2 to 3 (88) & .625 & .875 & +.250\\
Depth 4+ (57) & .649 & .789 & +.140\\
\bottomrule
\end{tabular}}
\captionof{table}{Recall by observability, state dependence, and interaction depth on the 160 Test faults.}
\label{tab:supp-blindspots}
\end{center}

\begin{center}
\centering
{\small
\setlength{\tabcolsep}{4pt}
\begin{tabular}{lrrrr}
\toprule
Fault family & $n$ & Original & Evolved & Gain\\
\midrule
Validation/auth & 24 & .583 & .875 & +.292\\
State/persistence & 50 & .620 & .860 & +.240\\
CRUD & 7 & .571 & .857 & +.286\\
Calculation & 16 & .625 & .813 & +.188\\
Search/filter/sort & 21 & .714 & .810 & +.096\\
Data binding & 19 & .684 & .789 & +.105\\
Cross-view & 23 & .739 & .913 & +.174\\
\bottomrule
\end{tabular}}
\captionof{table}{Fault recall by semantic family on Test.}
\label{tab:supp-families}
\end{center}

The strongest improvements occur in validation/authentication, CRUD, state persistence, and cross-view consistency, where correctness depends on a specific transition and its consequences. Search and data binding start from a stronger base and consequently have smaller absolute gains. Residual errors are most pronounced for depth-4+ flows and for data-binding and search interactions, whose final recall remains between 0.789 and 0.810.

\section{RL Protocol and Reward Audit}

\subsection{Optimization and Runtime}

\paragraph{SFT initialization.}
The RL warm start is checkpoint-7, the final checkpoint obtained after one epoch of full-parameter SFT on 600 rebuilt silver examples.
The queries are stratified samples from the public WebGen-Instruct train split.
Claude Sonnet 4.6 generates browser-executable Vite, React, and TypeScript targets in the WebGen-R1 artifact format.
Training packs the tokenized records into 210 fixed-length blocks and uses one epoch, a global batch size of 32, and a maximum sequence length of 16,384.
The optimizer is fused AdamW with learning rate $4\times10^{-5}$, $\beta_1=.9$, $\beta_2=.999$, $\epsilon=10^{-8}$, no weight decay, gradient clipping at 1.0, cosine decay, and 10\% warmup.
Training uses bf16 precision, gradient checkpointing, and ZeRO-3 without parameter or optimizer offload across eight GPUs.
The SFT seed is 3407.
The 101 WebGen-Bench test queries are excluded.

\begin{center}
\centering
{\small
\setlength{\tabcolsep}{3pt}
\begin{tabular}{lp{0.58\columnwidth}}
\toprule
Item & Frozen configuration\\
\midrule
Initialization & Qwen3-8B SFT checkpoint-7 (one epoch)\\
Training data & A different set of 600 queries independently sampled from the WebGen-Instruct train split\\
Framework & VERL + GRPO + vLLM rollout\\
Hardware & Single node, 8 $\times$ NVIDIA A100-SXM4 80GB\\
Effective updates & 100\\
Train batch & 4 queries per update\\
Group size & 8 completions per query (32 websites/update)\\
Learning rate & $2\times10^{-6}$\\
Prompt / response limit & 4,096 / 16,384 tokens\\
Rollout sampling & temperature .7, top-$p$ .95, top-$k$ 20\\
KL / entropy coefficient & .01 / 0\\
Checkpoint interval & 25 effective updates\\
Random seed & 20260707\\
\bottomrule
\end{tabular}}
\captionof{table}{Final WebGrader-RL hyperparameters.}
\label{tab:supp-rl-config}
\end{center}

All WebGrader-RL training statistics use the frozen run with seed 20260707, and evaluation tables use their fixed protocols and evaluator configurations.
GRPO uses AdamW with $\beta_1=.9$, $\beta_2=.999$, weight decay .01, a constant schedule without warmup, gradient clipping at 1.0, one PPO epoch, and clipping ratios .2/.2.
Advantages are normalized by the group standard deviation, and token-mean loss aggregation is used.
The actor and reference policies use FSDP in bf16 with parameter and optimizer offload, gradient checkpointing, SDPA attention, and a maximum of 24,576 tokens per GPU.

\begin{center}
\centering
{\small
\setlength{\tabcolsep}{3.2pt}
\begin{tabular}{lp{0.59\columnwidth}}
\toprule
Item & Reproduced environment\\
\midrule
Host & Ubuntu 20.04.5, Linux 5.4.250, x86-64\\
CPU / memory & 2 $\times$ Intel Xeon Platinum 8336C, 1.8 TiB host-visible RAM\\
GPU / driver & 8 $\times$ NVIDIA A100-SXM4 80GB, driver 535.129.03\\
CUDA & Toolkit 11.8 (V11.8.89)\\
RL stack & Python 3.12.13, PyTorch 2.10.0, Transformers 5.5.4, VERL 0.8.0.dev0, vLLM 0.19.1\\
SFT stack & Python 3.11.4, PyTorch 2.10.0, Transformers 5.2.0, DeepSpeed 0.18.8, Accelerate 1.11.0\\
Browser stack & Chromium 148.0.7729.0, Node.js 20.19.5, Playwright 1.55.0\\
\bottomrule
\end{tabular}}
\captionof{table}{Hardware and software environment. RAM is the host-visible amount because the container quota is not recoverable from the archived runtime.}
\label{tab:supp-environment}
\end{center}

Render and interaction viewports are 1280$\times$800 and 1280$\times$900.
The frozen WebGrader verifier requests DeepSeek-v4-flash with temperature 0 and thinking disabled.
Its Stage 1, Stage 2 router, Stage 2 program, Stage 3 base, and Stage 3 residual limits are 4,096, 3,000, 12,000, 1,200, and 1,800 output tokens.
The appearance model is GPT-5.4-mini with temperature 0 and a 512-token output limit.
Unspecified API parameters, including top-$p$ and seed, use provider defaults and are not retrospectively inferred.
For every completed rollout, the recorded scalar reward uses the same weighted sum as the main paper,
\[
R(q,x)=0.7\cdot 5I(q,x)+0.3A(x),
\]
where $I(q,x)\in[0,1]$ is the weighted functional score over determinate requirement-level flows and $A(x)\in[0,5]$ is the frozen appearance score.
No additional appearance gate or determinate-weight mask is applied.

\subsection{Training Data, Component Parity, and Evaluator Comparability}

The 100 WebGen-Verifier-100 queries, 600 SFT queries, and 600 RL queries were deliberately selected to be mutually disjoint at the requirement level.
The 600 RL queries form a separately sampled set and do not reuse the 600-query SFT set.
They are randomly sampled from the released WebGen-Instruct train split and never from the 101-instruction WebGen-Bench evaluation set.
The WebGen-Bench construction process first removes a candidate training instruction when its 5-gram Jaccard similarity to any evaluation instruction exceeds 0.6, then removes semantic near-duplicates whose all-MiniLM-L6-v2 sentence-embedding cosine similarity exceeds 0.55~\cite{lu2025webgenbench}.
The benchmark authors additionally inspect the three nearest remaining training instructions for every evaluation instruction.
Our RL subset preserves this released train/evaluation separation without admitting any WebGen-Bench evaluation instruction.

All generated websites reported on WebGen-Bench, including outputs from controlled policies and external generators, are evaluated from scratch with one frozen GPT-5.4 pipeline.
The GPT-5.4 evaluator version, appearance and UI-agent prompts, decoding settings, interaction limits, agent configuration, browser configuration, retry policy, and timeout policy are identical across generators.
The exact prompt and configuration artifacts and their content hashes are retained with the evaluation records.
No result in the comparison table is copied from an earlier evaluator version.
WG-core-250 is likewise run through one fixed benchmark pipeline for every generator.
Its deterministic browser cases and GPT-5.4 scoring configuration are unchanged across systems.

\begin{center}
\centering
{\small
\setlength{\tabcolsep}{3.5pt}
\begin{tabular}{lp{0.52\columnwidth}}
\toprule
Component & Controlled RL variants\\
\midrule
Initialization & Same Qwen3-8B SFT checkpoint-7\\
Training examples & Same 600 WebGen-Instruct train queries\\
Rollout budget & Same batches, group size, and 100 effective updates\\
Sampling & Same temperature, top-$p$, top-$k$, and token limits\\
Optimization & Same GRPO configuration, learning rate, and KL coefficient\\
Changed component & Reward construction only\\
\bottomrule
\end{tabular}}
\captionof{table}{Component parity for GUI-RL, VLM+GUI-RL, VLM+Base-Script-RL, and WebGrader-RL.}
\label{tab:supp-component-parity}
\end{center}

\subsection{Training-Time Verdict Audit and Aggregation Sensitivity}

Through effective update 100, the reward pipeline launched 3,712 attempts and returned 3,553 complete results containing 12,124 flow verdicts.
The raw Stage-3 output contains 2,159 \textsc{Inconclusive} verdicts.
Post-hoc attribution identifies 1,784 of these as website-caused \texttt{target\_not\_reached} outcomes because the required control or behavior is absent.
The reward semantics therefore count them deterministically as \textsc{Fail}.
The remaining 375 cases are evaluator-side uncertainty and are excluded from the weighted functional average.

\begin{center}
\centering
{\small
\setlength{\tabcolsep}{4pt}
\begin{tabular}{lrrrr}
\toprule
Verdict & Raw & Raw \% & Effective & Effective \%\\
\midrule
\textsc{Pass} & 5,930 & 48.91 & 5,930 & 48.91\\
\textsc{Fail} & 4,035 & 33.28 & 5,819 & 48.00\\
\textsc{Inconclusive} & 2,159 & 17.81 & 375 & 3.09\\
\bottomrule
\end{tabular}}
\captionof{table}{Training-time flow verdicts before and after applying the reward's site-side/evaluator-side attribution semantics.}
\label{tab:supp-verdict-audit}
\end{center}

As an offline sensitivity check, we recompute complete eight-sample groups after assigning zero functional credit to every remaining evaluator-side \textsc{Inconclusive} flow.
This intentionally stricter aggregation changes the mean recorded reward by $-0.0121$.
Although at least one pairwise relation changes in 19.21\% of groups, the selected top set changes in only 2.96\%, and the identity of a unique top completion changes in 1.97\%.

\begin{center}
\centering
{\small
\setlength{\tabcolsep}{5pt}
\begin{tabular}{lrr}
\toprule
Sensitivity statistic & Count & Rate\\
\midrule
Complete eight-sample groups & 406 & 100.00\%\\
Any pairwise relation changed & 78 & 19.21\%\\
Selected top set changed & 12 & 2.96\%\\
Unique top identity changed & 8 & 1.97\%\\
\bottomrule
\end{tabular}}
\captionof{table}{Offline sensitivity to replacing the remaining evaluator-side \textsc{Inconclusive} outcomes with zero functional credit.}
\label{tab:supp-inc-sensitivity}
\end{center}

\subsection{Execution and Model-Call Accounting}

\begin{center}
\centering
{\small
\setlength{\tabcolsep}{4.5pt}
\begin{tabular}{lr}
\toprule
Training cost item & Value\\
\midrule
Effective-update wall time & 5.61 h\\
Wall time incl. resampling & 6.13 h\\
Allocated GPU-hours & 49.07\\
Effective-path GPU-hours & 44.87\\
Reward jobs & 3,712\\
Mean completed reward job & 63.75 s\\
Build attempts & 3,527\\
Mean Vite build & 0.707 s\\
Render/probe executions & 3,303\\
Interactive flows & 12,124\\
Stage-2 cases & 3,186\\
Mean Stage-2 case & 37.54 s\\
\bottomrule
\end{tabular}}
\captionof{table}{Checkpoint-100 training and browser workload. Stage-2 time includes verifier script generation. Paused idle time is excluded.}
\label{tab:supp-cost}
\end{center}

The 3,712 attempts include retries and resampling around 3,200 policy completions. Of these, 3,553 returned complete reward records. Tables~\ref{tab:supp-cost} and~\ref{tab:supp-coverage} report the audited endpoints. Paused idle time is excluded, and the API usage totals below are lower bounds when failed requests return no usage record.

\paragraph{Browser execution coverage.}

\begin{center}
\centering
{\small
\setlength{\tabcolsep}{5pt}
\begin{tabular}{lrr}
\toprule
Outcome & Count & Rate\\
\midrule
Sampled completion & 3,200 & 100.00\%\\
Build success & 2,937 & 91.78\%\\
Build failure & 263 & 8.22\%\\
Render success & 2,928 & 91.50\%\\
Render failure & 272 & 8.50\%\\
Build success, render failure & 9 & 0.28\%\\
\bottomrule
\end{tabular}}
\captionof{table}{Browser execution coverage over 3,200 sampled completions. The nine build-success/render-failure cases separate compilation from browser launch.}
\label{tab:supp-coverage}
\end{center}

\paragraph{Model-call accounting.}

\begin{center}
\centering
{\small
\setlength{\tabcolsep}{7pt}
\begin{tabular}{lr}
\toprule
Stage & Successful calls\\
\midrule
Appearance & 3,037\\
Stage 1 + repair & 2,037\\
Stage 2 router + generation & 6,170\\
Stage 3 base + residual & 24,256\\
\midrule
Verifier subtotal & 32,463\\
All APIs & 35,500\\
\bottomrule
\end{tabular}
\par\smallskip
\setlength{\tabcolsep}{2.4pt}
\begin{tabular}{lrrr}
\toprule
Stage & Input & Output & Cached\\
\midrule
Appearance & 4,675,187 & 414,631 & 0\\
Stage 1 & 9,600,148 & 4,578,270 & 2,279,552\\
Stage 2 & 117,666,616 & 15,154,216 & 3,523,968\\
Stage 3 & 75,939,479 & 2,480,274 & 6,786,048\\
\midrule
Verifier & 203,206,243 & 22,212,760 & 12,589,568\\
All APIs & 207,881,430 & 22,627,391 & 12,589,568\\
\bottomrule
\end{tabular}}
\captionof{table}{Usage-bearing successful API requests through checkpoint 100. Requests that failed before returning usage may add a small amount, so these values are an auditable lower bound.}
\label{tab:supp-api}
\end{center}

\section{Complete Downstream Results}

\subsection{Complete WebGen-Bench Results}

\begin{center}
\centering
{\small
\setlength{\tabcolsep}{2.8pt}
\begin{tabular}{lrrrrrr}
\toprule
Method & VRR & AAS & Yes & Partial & No & FSR\\
\midrule
Qwen3-8B base & 90.10 & 1.51 & 70 & 53 & 524 & 14.92\\
WebGen SFT & 62.38 & 2.19 & 166 & 103 & 378 & 33.62\\
GUI-RL & 96.04 & 3.15 & 207 & 129 & 311 & 41.96\\
VLM+GUI-RL & 100.00 & 3.43 & 210 & 132 & 305 & 42.66\\
VLM+Base-Script-RL & 100.00 & 3.46 & 216 & 139 & 292 & 44.13\\
\textbf{WebGrader-RL} & \textbf{100.00} & \textbf{3.48} & \textbf{256} & \textbf{161} & \textbf{230} & \textbf{52.01}\\
\bottomrule
\end{tabular}}
\captionof{table}{Complete initialization-matched results on WebGen-Bench. The base-script comparison preserves the VLM appearance reward while removing verifier evolution.}
\label{tab:supp-webgen-complete}
\end{center}

For VLM+Base-Script-RL, the AAS grade distribution over the 101 pages is 2/9/38/45/7 for grades 1/2/3/4/5. The weighted total is 349, giving AAS $349/101=3.4554$. Its functional score is
\[
\frac{216+0.5(139)}{647}=44.1267\%.
\]
Relative to this baseline, WebGrader-RL adds 40 \textsc{Yes} and 22 \textsc{Partial} outcomes, removes 62 \textsc{No} outcomes, and improves FSR by 7.88 points.

\subsection{Functional Categories}

\begin{center}
\centering
{\small
\setlength{\tabcolsep}{3pt}
\begin{tabular}{lrrr}
\toprule
Category & Base script & WebGrader & Gain\\
\midrule
Content presentation & 54.31 & \textbf{63.80} & +9.49\\
User interaction & 41.69 & \textbf{48.20} & +6.51\\
Data management & 41.88 & \textbf{46.61} & +4.73\\
Functional testing & 30.68 & \textbf{36.00} & +5.32\\
Data display & 55.11 & \textbf{61.60} & +6.49\\
Design validation & 66.80 & \textbf{82.00} & +15.20\\
\bottomrule
\end{tabular}}
\captionof{table}{Direct category-level attribution on WebGen-Bench (FSR, \%).}
\label{tab:supp-category-attribution}
\end{center}

\begin{center}
\centering
{\small
\setlength{\tabcolsep}{2.2pt}
\begin{tabular}{lrrrr}
\toprule
Category & $n$ & Yes & Partial & No\\
\midrule
Content presentation & 174 & 75 & 39 & 60\\
User interaction & 313 & 102 & 57 & 154\\
Data management & 160 & 51 & 32 & 77\\
Functional testing & 339 & 78 & 52 & 209\\
Data display & 186 & 75 & 55 & 56\\
Design validation & 122 & 73 & 17 & 32\\
\bottomrule
\end{tabular}}
\captionof{table}{VLM+Base-Script-RL category outcome counts. Categories overlap, so their sizes do not sum to 647.}
\label{tab:supp-base-script-counts}
\end{center}

\subsection{External Model Context on WebGen-Bench}

\begin{center}
\centering
{\small
\setlength{\tabcolsep}{4pt}
\begin{tabular}{lrrr}
\toprule
Generator & VRR & AAS & FSR\\
\midrule
Claude Opus 4.8 & 100.00 & 3.96 & 68.20\\
GPT-5.5 & 90.10 & 3.78 & 62.44\\
Qwen3-Coder-480B & 94.06 & 3.32 & 53.25\\
GPT-5.4-mini & 89.11 & 3.27 & 52.32\\
\textbf{WebGrader-RL (8B)} & \textbf{100.00} & \textbf{3.48} & \textbf{52.01}\\
DeepSeek-v4-flash & 93.07 & 2.95 & 47.91\\
o4-mini & 89.11 & 2.49 & 40.65\\
\bottomrule
\end{tabular}}
\captionof{table}{External generator context under the same GPT-5.4 WebGen-Bench pipeline.}
\label{tab:supp-webgen-external}
\end{center}

\subsection{WG-core-250 Composition and Scoring}

WG-core-250 is a fixed 250-task subset of HTMLBench-400. It contains 50 applications, 55 UI components, 35 data-visualization pages, 35 content pages, 40 landing pages, and 35 portfolios. Every task contributes to the fixed denominator of 250. The HTMLBench Full score combines renderability (10), GPT-5.4 keyframe visual quality (20), functionality (55 for interactive tasks or 65 otherwise), interactivity (10 for interactive tasks or 0 otherwise), and code quality (5). TC\% is the deterministic browser test-case pass rate.

\begin{center}
\centering
{\small
\setlength{\tabcolsep}{3.4pt}
\begin{tabular}{lrrrr}
\toprule
Method & Score & TC\% & Rend. & E2E\\
\midrule
Qwen3-8B base & 31.160 & 27.418 & 6.004 & 171\\
WebGen SFT & 32.388 & 25.640 & 7.728 & 213\\
VLM+GUI-RL & 37.746 & 29.345 & 8.385 & 205\\
GUI-RL & 38.120 & 33.861 & 7.388 & 212\\
VLM+Base-Script-RL & 38.372 & 33.525 & 7.492 & 210\\
o4-mini & 39.068 & 31.668 & 8.648 & 229\\
Qwen3-Coder-480B & 42.504 & 36.328 & 8.420 & 222\\
DeepSeek-v4-flash & 43.684 & 36.103 & 9.016 & 242\\
\textbf{WebGrader-RL} & \textbf{44.953} & \textbf{39.931} & 8.712 & 212\\
GPT-5.4-mini & 49.636 & 42.277 & 9.656 & 250\\
Claude Opus 4.8 & 50.460 & 45.653 & 9.192 & 239\\
GPT-5.5 & 52.280 & 47.783 & 9.356 & 242\\
\bottomrule
\end{tabular}}
\captionof{table}{WG-core-250 aggregate and rendering results. E2E reports rendered outputs out of 250, while aggregate scores retain the fixed denominator.}
\label{tab:supp-wgcore}
\end{center}

\begin{center}
\centering
{\small
\setlength{\tabcolsep}{3.4pt}
\begin{tabular}{lrrrr}
\toprule
Method & Vis. & Func. & Inter. & Code\\
\midrule
Qwen3-8B base & 5.808 & 15.644 & .968 & 2.736\\
WebGen SFT & 5.540 & 14.540 & 1.184 & 3.396\\
VLM+GUI-RL & 7.185 & 16.605 & 1.493 & 4.078\\
GUI-RL & 6.928 & 19.088 & 1.324 & 3.392\\
VLM+Base-Script-RL & 7.148 & 18.920 & 1.452 & 3.360\\
o4-mini & 7.704 & 17.876 & 1.176 & 3.664\\
Qwen3-Coder-480B & 8.556 & 20.332 & 1.648 & 3.548\\
DeepSeek-v4-flash & 8.876 & 20.368 & 1.556 & 3.868\\
\textbf{WebGrader-RL} & 8.170 & \textbf{22.509} & 1.561 & 4.000\\
GPT-5.4-mini & 10.288 & 23.844 & 1.848 & 4.000\\
Claude Opus 4.8 & 10.044 & 25.680 & 1.720 & 3.824\\
GPT-5.5 & 10.316 & 26.868 & 1.868 & 3.872\\
\bottomrule
\end{tabular}}
\captionof{table}{WG-core-250 visual, functional, interactive, and code components.}
\label{tab:supp-wgcore-components}
\end{center}

WebGrader-RL improves the matched VLM+Base-Script-RL policy by 6.581 points, the 8B base by 13.793, WebGen SFT by 12.565, GUI-RL by 6.833, and VLM+GUI-RL by 7.207. Against external generators, it exceeds o4-mini by 5.885, Qwen3-Coder-480B by 2.449, and DeepSeek-v4-flash by 1.269. Its functional component is 22.509, the strongest among the models below the top three proprietary references.

\subsection{Paired Statistical Analysis}

We complement the official aggregate metrics with paired query- and task-level tests.
For WebGen-Bench, each of the 101 website queries is one resampling unit.
Within a query, \textsc{Yes}, \textsc{Partial}, and \textsc{No} receive 1, .5, and 0 credit, respectively, and are averaged before taking the macro mean across queries.
This query-macro statistic gives every website equal weight, whereas the official FSR in Table~\ref{tab:supp-webgen-complete} is the micro average over 647 functional cases.
Confidence intervals use 10,000 paired bootstrap resamples of queries.
The two-sided paired permutation reference uses 10,000 random sign permutations, and the Wilcoxon test operates on the 101 paired query scores.

\begin{center}
\centering
{\footnotesize
\setlength{\tabcolsep}{2pt}
\begin{tabular}{lrr}
\toprule
Query-macro statistic & Result & Paired inference\\
\midrule
VLM+Base-Script-RL FSR & 46.3\% & \\
WebGrader-RL FSR & 52.0\% & \\
Difference & +5.7 pp & 95\% CI $[+1.9,+9.5]$ pp\\
\midrule
Permutation test & & $p=.004$\\
Wilcoxon signed-rank & & $p=.006$\\
\bottomrule
\end{tabular}}
\captionof{table}{Paired query-level analysis on WebGen-Bench.}
\label{tab:supp-webgen-significance}
\end{center}

For WG-core-250, each benchmark task is one paired unit.
The evaluator's normalized task-level Full Score and per-task test-case ratio are averaged across the 250 tasks for this analysis.
These task-macro diagnostic means are reported separately from the fixed-denominator aggregate Score and TC\% in Table~\ref{tab:supp-wgcore}.
Bootstrap confidence intervals use 10,000 paired resamples of tasks.

\begin{center}
\centering
{\scriptsize
\setlength{\tabcolsep}{1.8pt}
\begin{tabular}{lrrr}
\toprule
Task-macro metric & Base & WebGrader & Delta [95\% CI]\\
\midrule
Normalized Full Score & 61.6\% & 72.8\% & $+11.2$ pp $[+6.0,+16.4]$\\
Test-case ratio & 71.3\% & 80.1\% & $+8.8$ pp $[+6.1,+11.6]$\\
\bottomrule
\end{tabular}}
\captionof{table}{Paired task-level analysis on WG-core-250. The Full Score difference has Wilcoxon $p<.001$.}
\label{tab:supp-wgcore-significance}
\end{center}

For a binary browser-test reference, a task is correct when the evaluator returns a positive task-level browser-test outcome.
The matched policies are jointly correct on 145 tasks and jointly incorrect on 49.
WebGrader alone is correct on 43 tasks, compared with 13 for the base-script policy.
The resulting exact two-sided McNemar test gives $p=7.3\times10^{-5}$.
The marginal counts are 158/250 for VLM+Base-Script-RL and 188/250 for WebGrader-RL.

\ifdefined\webgraderappendix\else
\bibliography{references}

\begin{thebibliography}{33}
\providecommand{\natexlab}[1]{#1}

\bibitem[{Austin et~al.(2021)Austin, Odena, Nye, Bosma, Michalewski, Dohan,
  Jiang, Cai, Terry, Le, and Sutton}]{austin2021mbpp}
Austin, J.; Odena, A.; Nye, M.; Bosma, M.; Michalewski, H.; Dohan, D.; Jiang,
  E.; Cai, C.; Terry, M.; Le, Q.; and Sutton, C. 2021.
\newblock Program Synthesis with Large Language Models.
\newblock \emph{arXiv preprint arXiv:2108.07732}.

\bibitem[{Chen et~al.(2021)Chen, Tworek, Jun, Yuan, Pinto, Kaplan, Edwards,
  Burda, Joseph, Brockman et~al.}]{chen2021humaneval}
Chen, M.; Tworek, J.; Jun, H.; Yuan, Q.; Pinto, H. P. d.~O.; Kaplan, J.;
  Edwards, H.; Burda, Y.; Joseph, N.; Brockman, G.; et~al. 2021.
\newblock Evaluating Large Language Models Trained on Code.
\newblock \emph{arXiv preprint arXiv:2107.03374}.

\bibitem[{Drouin et~al.(2024)Drouin, Gasse, Caccia, Laradji, Del~Verme, Marty,
  Vazquez, Chapados, and Lacoste}]{drouin2024workarena}
Drouin, A.; Gasse, M.; Caccia, M.; Laradji, I.~H.; Del~Verme, M.; Marty, T.;
  Vazquez, D.; Chapados, N.; and Lacoste, A. 2024.
\newblock {WorkArena}: How Capable Are Web Agents at Solving Common Knowledge
  Work Tasks?
\newblock In \emph{Proceedings of the 41st International Conference on Machine
  Learning}, volume 235 of \emph{Proceedings of Machine Learning Research},
  11642--11662. PMLR.

\bibitem[{Gao et~al.(2026)Gao, Chen, He, Wang, Xu, Wang, Jin, and
  Wu}]{gao2026eigendata}
Gao, J.; Chen, J.; He, C.; Wang, W.-C.; Xu, S.; Wang, H.; Jin, D.; and Wu, Y.
  2026.
\newblock From Self-Evolving Synthetic Data to Verifiable-Reward {RL}:
  Post-Training Multi-Turn Interactive Tool-Using Agents.
\newblock \emph{arXiv preprint arXiv:2601.22607}.

\bibitem[{Gunjal et~al.(2026)Gunjal, Wang, Lau, Nath, He, Liu, and
  Hendryx}]{gunjal2026rubrics}
Gunjal, A.; Wang, A.; Lau, E.; Nath, V.; He, Y.; Liu, B.; and Hendryx, S. 2026.
\newblock Rubrics as Rewards: Reinforcement Learning Beyond Verifiable Domains.
\newblock In \emph{The Fourteenth International Conference on Learning
  Representations}.

\bibitem[{He et~al.(2026)He, Weir, Bostrom, Nie, Cassel, Bayless, and
  Rangwala}]{he2026resyn}
He, A.; Weir, N.; Bostrom, K.; Nie, A.; Cassel, D.; Bayless, S.; and Rangwala,
  H. 2026.
\newblock {ReSyn}: Autonomously Scaling Synthetic Environments for Reasoning
  Models.
\newblock \emph{arXiv preprint arXiv:2602.20117}.

\bibitem[{He et~al.(2024)He, Yao, Ma, Yu, Dai, Zhang, Lan, and
  Yu}]{he2024webvoyager}
He, H.; Yao, W.; Ma, K.; Yu, W.; Dai, Y.; Zhang, H.; Lan, Z.; and Yu, D. 2024.
\newblock {WebVoyager}: Building an End-to-End Web Agent with Large Multimodal
  Models.
\newblock In \emph{Proceedings of the 62nd Annual Meeting of the Association
  for Computational Linguistics (Volume 1: Long Papers)}, 6864--6890. Bangkok,
  Thailand: Association for Computational Linguistics.

\bibitem[{Hendrycks et~al.(2021)Hendrycks, Basart, Kadavath, Mazeika, Arora,
  Guo, Burns, Puranik, He, Song, and Steinhardt}]{hendrycks2021apps}
Hendrycks, D.; Basart, S.; Kadavath, S.; Mazeika, M.; Arora, A.; Guo, E.;
  Burns, C.; Puranik, S.; He, H.; Song, D.; and Steinhardt, J. 2021.
\newblock Measuring Coding Challenge Competence with {APPS}.
\newblock In \emph{Proceedings of the Neural Information Processing Systems
  Track on Datasets and Benchmarks}, volume~1.

\bibitem[{Ji et~al.(2026)Ji, Yang, Song, Wang, Cui, Li, Jiang, and
  Chen}]{ji2026finestate}
Ji, F.; Yang, J.; Song, Z.; Wang, Y.; Cui, Z.; Li, Y.; Jiang, Q.; and Chen, X.
  2026.
\newblock {FineState-Bench}: Benchmarking State-Conditioned Grounding for
  Fine-Grained {GUI} State Setting.
\newblock In \emph{Findings of the Association for Computational Linguistics:
  ACL 2026}, 43073--43088. San Diego, California, United States: Association
  for Computational Linguistics.

\bibitem[{Jiang et~al.(2026)Jiang, Cai, Park, Shen, Kim, Li, and
  Wang}]{jiang2026webgenr1}
Jiang, J.; Cai, C.; Park, C.; Shen, J.; Kim, S.; Li, J.; and Wang, Y. 2026.
\newblock {WebGen-R1}: Incentivizing Large Language Models to Generate
  Functional and Aesthetic Websites with Reinforcement Learning.
\newblock \emph{arXiv preprint arXiv:2604.20398}.

\bibitem[{Jimenez et~al.(2024)Jimenez, Yang, Wettig, Yao, Pei, Press, and
  Narasimhan}]{jimenez2023swebench}
Jimenez, C.~E.; Yang, J.; Wettig, A.; Yao, S.; Pei, K.; Press, O.; and
  Narasimhan, K. 2024.
\newblock {SWE-bench}: Can Language Models Resolve Real-World {GitHub} Issues?
\newblock In \emph{The Twelfth International Conference on Learning
  Representations}.

\bibitem[{Kim et~al.(2024)Kim, Shin, Cho, Jang, Longpre, Lee, Yun, Shin, Kim,
  Thorne, and Seo}]{kim2023prometheus}
Kim, S.; Shin, J.; Cho, Y.; Jang, J.; Longpre, S.; Lee, H.; Yun, S.; Shin, S.;
  Kim, S.; Thorne, J.; and Seo, M. 2024.
\newblock {Prometheus}: Inducing Fine-Grained Evaluation Capability in Language
  Models.
\newblock In \emph{The Twelfth International Conference on Learning
  Representations}.

\bibitem[{Koh et~al.(2024)Koh, Lo, Jang, Duvvur, Lim, Huang, Neubig, Zhou,
  Salakhutdinov, and Fried}]{koh2024visualwebarena}
Koh, J.~Y.; Lo, R.; Jang, L.; Duvvur, V.; Lim, M.; Huang, P.-Y.; Neubig, G.;
  Zhou, S.; Salakhutdinov, R.; and Fried, D. 2024.
\newblock {VisualWebArena}: Evaluating Multimodal Agents on Realistic Visual
  Web Tasks.
\newblock In \emph{Proceedings of the 62nd Annual Meeting of the Association
  for Computational Linguistics (Volume 1: Long Papers)}, 881--905. Bangkok,
  Thailand: Association for Computational Linguistics.

\bibitem[{Kong et~al.(2026)Kong, Zhang, Yue, Sun, Tian, Feng, Yang, Wang, Tian,
  Du, Zeng, Li, and Gai}]{kong2026webtestbench}
Kong, F.; Zhang, J.; Yue, Y.; Sun, C.; Tian, Y.; Feng, S.; Yang, X.; Wang, D.;
  Tian, Y.; Du, J.; Zeng, W.; Li, H.; and Gai, K. 2026.
\newblock {WebTestBench}: Evaluating Computer-Use Agents towards End-to-End
  Automated Web Testing.
\newblock \emph{arXiv preprint arXiv:2603.25226}.

\bibitem[{Lai et~al.(2025)Lai, Zhuang, Zhang, Xiong, Wang, Xu, Chen, Wang, and
  Cui}]{lai2025webrenderbench}
Lai, P.; Zhuang, J.; Zhang, K.; Xiong, N.; Wang, S.; Xu, Y.; Chen, C.; Wang,
  Y.; and Cui, B. 2025.
\newblock {WebRenderBench}: Enhancing Web Interface Generation through
  Layout-Style Consistency and Reinforcement Learning.
\newblock \emph{arXiv preprint arXiv:2510.04097}.

\bibitem[{Li et~al.(2026)Li, Zhang, Lv, Liu, Deng, Zhang, Liu, and
  Zhou}]{li2025relook}
Li, Y.; Zhang, C.; Lv, R.; Liu, A.; Deng, K.; Zhang, Y.; Liu, J.; and Zhou, B.
  2026.
\newblock {ReLook}: Vision-Grounded {RL} with a Multimodal {LLM} Critic for
  Agentic Web Coding.
\newblock In \emph{Proceedings of the 64th Annual Meeting of the Association
  for Computational Linguistics (Volume 1: Long Papers)}, 25471--25485. San
  Diego, California, United States: Association for Computational Linguistics.

\bibitem[{Liu et~al.(2023{\natexlab{a}})Liu, Xia, Wang, and
  Zhang}]{liu2023evalplus}
Liu, J.; Xia, C.~S.; Wang, Y.; and Zhang, L. 2023{\natexlab{a}}.
\newblock Is Your Code Generated by {ChatGPT} Really Correct? Rigorous
  Evaluation of Large Language Models for Code Generation.
\newblock In \emph{Advances in Neural Information Processing Systems},
  volume~36, 21558--21572. Curran Associates, Inc.

\bibitem[{Liu et~al.(2023{\natexlab{b}})Liu, Iter, Xu, Wang, Xu, and
  Zhu}]{liu2023geval}
Liu, Y.; Iter, D.; Xu, Y.; Wang, S.; Xu, R.; and Zhu, C. 2023{\natexlab{b}}.
\newblock {G-Eval}: {NLG} Evaluation Using {GPT-4} with Better Human Alignment.
\newblock In \emph{Proceedings of the 2023 Conference on Empirical Methods in
  Natural Language Processing}, 2511--2522. Singapore: Association for
  Computational Linguistics.

\bibitem[{Lu et~al.(2026)Lu, Ren, Yang, Wang, Zong, Pan, Zhan, and
  Li}]{lu2026webgenagent}
Lu, Z.; Ren, H.; Yang, Y.; Wang, K.; Zong, Z.; Pan, J.; Zhan, M.; and Li, H.
  2026.
\newblock {WebGen-Agent}: Enhancing Interactive Website Generation with
  Multi-Level Feedback and Step-Level Reinforcement Learning.
\newblock In \emph{The Fourteenth International Conference on Learning
  Representations}.

\bibitem[{Lu et~al.(2025)Lu, Yang, Ren, Hou, Xiao, Wang, Shi, Zhou, Zhan, and
  Li}]{lu2025webgenbench}
Lu, Z.; Yang, Y.; Ren, H.; Hou, H.; Xiao, H.; Wang, K.; Shi, W.; Zhou, A.;
  Zhan, M.; and Li, H. 2025.
\newblock {WebGen-Bench}: Evaluating {LLM}s on Generating Interactive and
  Functional Websites from Scratch.
\newblock In \emph{Advances in Neural Information Processing Systems},
  volume~38. Curran Associates, Inc.

\bibitem[{Peng et~al.(2025)Peng, Qi, Wang, Xu, Hou, and Li}]{peng2025verif}
Peng, H.; Qi, Y.; Wang, X.; Xu, B.; Hou, L.; and Li, J. 2025.
\newblock {VerIF}: Verification Engineering for Reinforcement Learning in
  Instruction Following.
\newblock In \emph{Proceedings of the 2025 Conference on Empirical Methods in
  Natural Language Processing}, 30324--30339. Suzhou, China: Association for
  Computational Linguistics.

\bibitem[{Peng et~al.(2026)Peng, Tao, Yin, Ying, Luo, and
  Guo}]{peng2026playcoder}
Peng, Z.; Tao, W.; Yin, X.; Ying, C.; Luo, Y.; and Guo, Y. 2026.
\newblock {PlayCoder}: Making {LLM}-Generated {GUI} Code Playable.
\newblock \emph{Proceedings of the ACM on Software Engineering}, 3(FSE):
  2003--2026.

\bibitem[{Ruan et~al.(2026)Ruan, Jiang, Zeng, Nie, and
  Chen}]{ruan2026evolvecoder}
Ruan, C.; Jiang, D.; Zeng, H.; Nie, P.; and Chen, W. 2026.
\newblock {EvolveCoder}: Evolving Test Cases via Adversarial Verification for
  Code Reinforcement Learning.
\newblock \emph{arXiv preprint arXiv:2603.12698}.

\bibitem[{Shao et~al.(2024)Shao, Wang, Zhu, Xu, Song, Bi, Zhang, Zhang, Li, Wu,
  and Guo}]{shao2024deepseekmath}
Shao, Z.; Wang, P.; Zhu, Q.; Xu, R.; Song, J.; Bi, X.; Zhang, H.; Zhang, M.;
  Li, Y.~K.; Wu, Y.; and Guo, D. 2024.
\newblock {DeepSeekMath}: Pushing the Limits of Mathematical Reasoning in Open
  Language Models.
\newblock \emph{arXiv preprint arXiv:2402.03300}.

\bibitem[{Sheng et~al.(2025)Sheng, Zhang, Ye, Wu, Zhang, Zhang, Peng, Lin, and
  Wu}]{sheng2024hybridflow}
Sheng, G.; Zhang, C.; Ye, Z.; Wu, X.; Zhang, W.; Zhang, R.; Peng, Y.; Lin, H.;
  and Wu, C. 2025.
\newblock {HybridFlow}: A Flexible and Efficient {RLHF} Framework.
\newblock In \emph{Proceedings of the Twentieth European Conference on Computer
  Systems}, EuroSys '25, 1279--1297. Association for Computing Machinery.

\bibitem[{Si et~al.(2025)Si, Zhang, Li, Yang, Liu, and
  Yang}]{si2024design2code}
Si, C.; Zhang, Y.; Li, R.; Yang, Z.; Liu, R.; and Yang, D. 2025.
\newblock {Design2Code}: Benchmarking Multimodal Code Generation for Automated
  Front-End Engineering.
\newblock In \emph{Proceedings of the 2025 Conference of the Nations of the
  Americas Chapter of the Association for Computational Linguistics: Human
  Language Technologies (Volume 1: Long Papers)}, 3956--3974. Albuquerque, New
  Mexico: Association for Computational Linguistics.

\bibitem[{Wu et~al.(2026)Wu, Yang, Zheng, Zhang, Wang, Lou, and
  Liu}]{wu2026htmlcure}
Wu, J.; Yang, J.; Zheng, T.; Zhang, W.; Wang, H.; Lou, Y.; and Liu, X. 2026.
\newblock {HTMLCure}: Turning Browser Experience into State Guided Repair for
  Interactive {HTML}.
\newblock \emph{arXiv preprint arXiv:2605.26807}.

\bibitem[{Yang et~al.(2025)Yang, Li, Yang, Zhang, Hui, Zheng, Yu, Gao, Huang,
  Lv, Zheng, Liu, Zhou, Huang, Hu, Ge, Wei, Lin, Tang, Yang, Tu, Zhang, Yang,
  Yang, Zhou, Zhou, Lin, Dang, Bao, Yang, Yu, Deng, Li, Xue, Li, Zhang, Wang,
  Zhu, Men, Gao, Liu, Luo, Li, Tang, Yin, Ren, Wang, Zhang, Ren, Fan, Su,
  Zhang, Zhang, Wan, Liu, Wang, Cui, Zhang, Zhou, and Qiu}]{yang2025qwen3}
Yang, A.; Li, A.; Yang, B.; Zhang, B.; Hui, B.; Zheng, B.; Yu, B.; Gao, C.;
  Huang, C.; Lv, C.; Zheng, C.; Liu, D.; Zhou, F.; Huang, F.; Hu, F.; Ge, H.;
  Wei, H.; Lin, H.; Tang, J.; Yang, J.; Tu, J.; Zhang, J.; Yang, J.; Yang, J.;
  Zhou, J.; Zhou, J.; Lin, J.; Dang, K.; Bao, K.; Yang, K.; Yu, L.; Deng, L.;
  Li, M.; Xue, M.; Li, M.; Zhang, P.; Wang, P.; Zhu, Q.; Men, R.; Gao, R.; Liu,
  S.; Luo, S.; Li, T.; Tang, T.; Yin, W.; Ren, X.; Wang, X.; Zhang, X.; Ren,
  X.; Fan, Y.; Su, Y.; Zhang, Y.; Zhang, Y.; Wan, Y.; Liu, Y.; Wang, Z.; Cui,
  Z.; Zhang, Z.; Zhou, Z.; and Qiu, Z. 2025.
\newblock {Qwen3} Technical Report.
\newblock \emph{arXiv preprint arXiv:2505.09388}.

\bibitem[{Zhang et~al.(2026)Zhang, Wang, Zhang, Guo, Li, Li, and
  Lu}]{zhang2026infiniteweb}
Zhang, Z.; Wang, Z.; Zhang, X.; Guo, Z.; Li, J.; Li, B.; and Lu, Y. 2026.
\newblock {InfiniteWeb}: Scalable Web Environment Synthesis for {GUI} Agent
  Training.
\newblock In \emph{Proceedings of the 64th Annual Meeting of the Association
  for Computational Linguistics (Volume 1: Long Papers)}, 28465--28492. San
  Diego, California, United States: Association for Computational Linguistics.

\bibitem[{Zheng et~al.(2024)Zheng, Gou, Kil, Sun, and Su}]{zheng2024seeact}
Zheng, B.; Gou, B.; Kil, J.; Sun, H.; and Su, Y. 2024.
\newblock {GPT-4V}(ision) Is a Generalist Web Agent, If Grounded.
\newblock In \emph{Proceedings of the 41st International Conference on Machine
  Learning}, volume 235 of \emph{Proceedings of Machine Learning Research},
  61349--61385. PMLR.

\bibitem[{Zheng et~al.(2023)Zheng, Chiang, Sheng, Zhuang, Wu, Zhuang, Lin, Li,
  Li, Xing, Zhang, Gonzalez, and Stoica}]{zheng2023llmjudge}
Zheng, L.; Chiang, W.-L.; Sheng, Y.; Zhuang, S.; Wu, Z.; Zhuang, Y.; Lin, Z.;
  Li, Z.; Li, D.; Xing, E.~P.; Zhang, H.; Gonzalez, J.~E.; and Stoica, I. 2023.
\newblock Judging {LLM}-as-a-Judge with {MT-Bench} and Chatbot Arena.
\newblock In \emph{Advances in Neural Information Processing Systems},
  volume~36, 46595--46623. Curran Associates, Inc.

\bibitem[{Zhou et~al.(2024)Zhou, Xu, Zhu, Zhou, Lo, Sridhar, Cheng, Ou, Bisk,
  Fried, Alon, and Neubig}]{zhou2023webarena}
Zhou, S.; Xu, F.~F.; Zhu, H.; Zhou, X.; Lo, R.; Sridhar, A.; Cheng, X.; Ou, T.;
  Bisk, Y.; Fried, D.; Alon, U.; and Neubig, G. 2024.
\newblock {WebArena}: A Realistic Web Environment for Building Autonomous
  Agents.
\newblock In \emph{The Twelfth International Conference on Learning
  Representations}.

\bibitem[{Zoph and Le(2017)}]{zoph2017nas}
Zoph, B.; and Le, Q.~V. 2017.
\newblock Neural Architecture Search with Reinforcement Learning.
\newblock In \emph{International Conference on Learning Representations}.

\end{thebibliography}
\end{document}
\fi

\end{document}